\documentclass{article} %
\usepackage{iclr2024_conference,times}

\usepackage{color,xcolor}
\usepackage{epsfig}
\usepackage{graphicx}

\usepackage{adjustbox}
\usepackage{array}
\usepackage{booktabs}
\usepackage{colortbl}
\usepackage{float,wrapfig}
\usepackage{hhline}
\usepackage{multirow}
\usepackage{subcaption} %
\usepackage[font={small}]{caption}
\usepackage{natbib}

\usepackage{amsmath,amsfonts,amsthm,amssymb}
\usepackage{bm}
\usepackage{nicefrac}
\usepackage{microtype}
\usepackage{mathtools}

\usepackage{changepage}
\usepackage{extramarks}
\usepackage{fancyhdr}
\usepackage{lastpage}
\usepackage{setspace}
\usepackage{soul}
\usepackage{xspace}

\usepackage{url}

\usepackage{algorithmic}
\usepackage{algorithm}
\usepackage{todonotes} %
\usepackage{enumitem}
\usepackage{titlesec}
\usepackage{bbm}

\usepackage[pagebackref=true,breaklinks=true,colorlinks,citecolor=gray]{hyperref}

\hypersetup{
    colorlinks=true,
    linkcolor=orange,
    filecolor=magenta,      
    urlcolor=magenta,
    citecolor=[rgb]{0,0.07843,0.45098}
}
\newcommand{\model}{VLP\xspace}
\newcommand{\sect}[1]{Sec.~\ref{#1}}

\newcommand{\eqn}[1]{Eqn~(\ref{#1})}
\newcommand{\fig}[1]{Fig.~\ref{#1}}
\newcommand{\tbl}[1]{Tab.~\ref{#1}}

\newcommand{\myparagraph}[1]{{\bf #1}\quad}

\newcommand{\ie}{i.e., }
\newcommand{\eg}{e.g., }

\usepackage{amsmath,amsfonts,bm}

\def\eqref#1{equation~\ref{#1}}

\def\1{\bm{1}}

\DeclareMathAlphabet{\mathsfit}{\encodingdefault}{\sfdefault}{m}{sl}
\SetMathAlphabet{\mathsfit}{bold}{\encodingdefault}{\sfdefault}{bx}{n}

\DeclareMathOperator*{\argmax}{arg\,max}

\usepackage{hyperref}
\usepackage{url}
\usepackage{tikz}
\usetikzlibrary{positioning}
\usetikzlibrary{decorations.pathmorphing}
\tikzset{snake it/.style={decorate, decoration=snake}}
\usepackage{ctable}
\usepackage{subcaption}

\makeatletter
\newcommand{\printfnsymbol}[1]{%
  \textsuperscript{\@fnsymbol{#1}}%
}
\makeatother

\title{Video Language Planning}

\iclrfinalcopy

\author{
Yilun Du\printfnsymbol{2}\printfnsymbol{3}, ~Mengjiao Yang\printfnsymbol{2}\printfnsymbol{4}, ~Pete Florence\printfnsymbol{2}, ~Fei Xia\printfnsymbol{2}, ~Ayzaan Wahid\printfnsymbol{2}, ~Brian Ichter\printfnsymbol{2},\\
~\textbf{Pierre Sermanet}\printfnsymbol{2}, ~\textbf{Tianhe Yu}\printfnsymbol{2},  ~\textbf{Pieter Abbeel}\printfnsymbol{4},  ~\textbf{Joshua B.  Tenenbaum}\printfnsymbol{3},  ~\textbf{Leslie Kaelbling}\printfnsymbol{3} \\
~\textbf{Andy Zeng}\printfnsymbol{2}, ~\textbf{Jonathan Tompson }\printfnsymbol{2} \\
~Google Deepmind\printfnsymbol{2},
~Massachusetts Institute of Technology\printfnsymbol{3},
~UC Berkeley\printfnsymbol{4}\\
\tt\href{https://video-language-planning.github.io/}{https://video-language-planning.github.io/}\\
}

\begin{document}

\maketitle

\setlength{\abovedisplayskip}{2pt}
\setlength{\abovedisplayshortskip}{2pt}
\setlength{\belowdisplayskip}{2pt}
\setlength{\belowdisplayshortskip}{2pt}
\setlength{\jot}{1pt}
\setlength{\textfloatsep}{1.5ex}
\setlength{\parskip}{0.38em}
\titlespacing\section{2pt}{6pt plus 1pt minus 1pt}{5pt plus 1pt minus 1pt}
\titlespacing\subsection{2pt}{5pt plus 1pt minus 1pt}{5pt plus 1pt minus 1pt}
\makeatletter
\renewcommand{\paragraph}{%
  \@startsection{paragraph}{4}%
  {\z@}{0.05ex \@plus .05ex \@minus .05ex}{-1em}%
  {\normalfont\normalsize\bfseries}%
}
\setlength{\floatsep}{2ex}

\begin{abstract}
We are interested in enabling visual planning for complex long-horizon tasks in the space of generated videos and language, leveraging recent advances in large generative models pretrained on Internet-scale data. 
To this end, we present \textit{video language planning} (VLP), an algorithm that consists of a tree search procedure, where we train (i) vision-language models to serve as both policies and value functions, and (ii) text-to-video models as dynamics models.
VLP takes as input a long-horizon task instruction and current image observation, and outputs a long video plan that provides detailed multimodal (video and language) specifications that describe how to complete the final task.
VLP scales with increasing computation budget where more computation time results in improved video plans,
and is able to synthesize long-horizon video plans across different robotics domains -- from multi-object rearrangement, to multi-camera bi-arm dexterous manipulation.
Generated video plans can be translated into real robot actions via goal-conditioned policies, conditioned on each intermediate frame of the generated video. Experiments show that VLP substantially improves long-horizon task success rates compared to prior methods on both simulated and real robots (across 3 hardware platforms).
\end{abstract}
\section{Introduction}

\looseness=-1
Intelligently interacting with the physical world involves planning over both (i) high-level semantic abstractions of the task (\ie what to do next), as well as the (ii) low-level underlying dynamics of the world (\ie how the world works). Factorizing the planning problem into two parts, one driven by task-specific objectives and the other a task-agnostic modeling of state transitions, is an idea that is pervasive and fundamental. This factorization drives much of the classic work in robotics from integrating task and motion planning \citep{cambon2009hybrid,wolfe2010combined,kaelbling2011hierarchical} to deriving control policies that can perform complex manipulation tasks over long time horizons such as tidying a dining table or rearranging a collection of objects to build new structures.

Pre-trained large language models (LLMs) \citep{brown2020language,chowdhery2022palm} have shown to be capable of generating high-level step-by-step plans for long-horizon tasks
over symbolic (often linguistic) abstractions of the task \citep{huang2022language,ahn2022can}, but this is only part of the solution. LLMs are restricted by what they can represent in text, and struggle with grounding \ie reasoning over shapes, physics, and constraints of the real world \citep{tellex2020robots,huang2023grounded}. LLMs can be integrated into larger vision-language models (VLMs) \citep{driess2023palm} that, when trained on sufficient data, can respect physical constraints observed in image inputs to generate more feasible plans that may be less likely to command the robot to perform impossible actions, or manipulate inaccessible objects. However, existing VLMs are predominantly trained on static image captioning and Q\&A datasets -- consequently, they continue to struggle to reason over dynamics \eg how objects may move or collide with one another over time. 

Meanwhile, recent text-to-video models trained on the wealth of videos on the Internet \citep{villegas2022phenaki,ho2022imagen}, have demonstrated an ability to learn the dynamics and motions of objects by synthesizing detailed video predictions of the future \citep{du2023learning}. Existing video models can only generate short time horizon clips without losing visual fidelity, and whether they can be applied for long-horizon planning remains unclear. Nevertheless, they exhibit properties that are complementary to VLMs in that they (i) can model the low-level visual dynamics of objects in ways that are more information-rich than text, and (ii) can absorb another source of Internet data \eg YouTube videos. 
This leads to the natural question of how to build a planning algorithm that can leverage both long-horizon abstract planning from LLMs / VLMs and detailed dynamics and motions from text-to-video-models.

\begin{figure}[t]
\begin{center}
    \includegraphics[width=\linewidth]{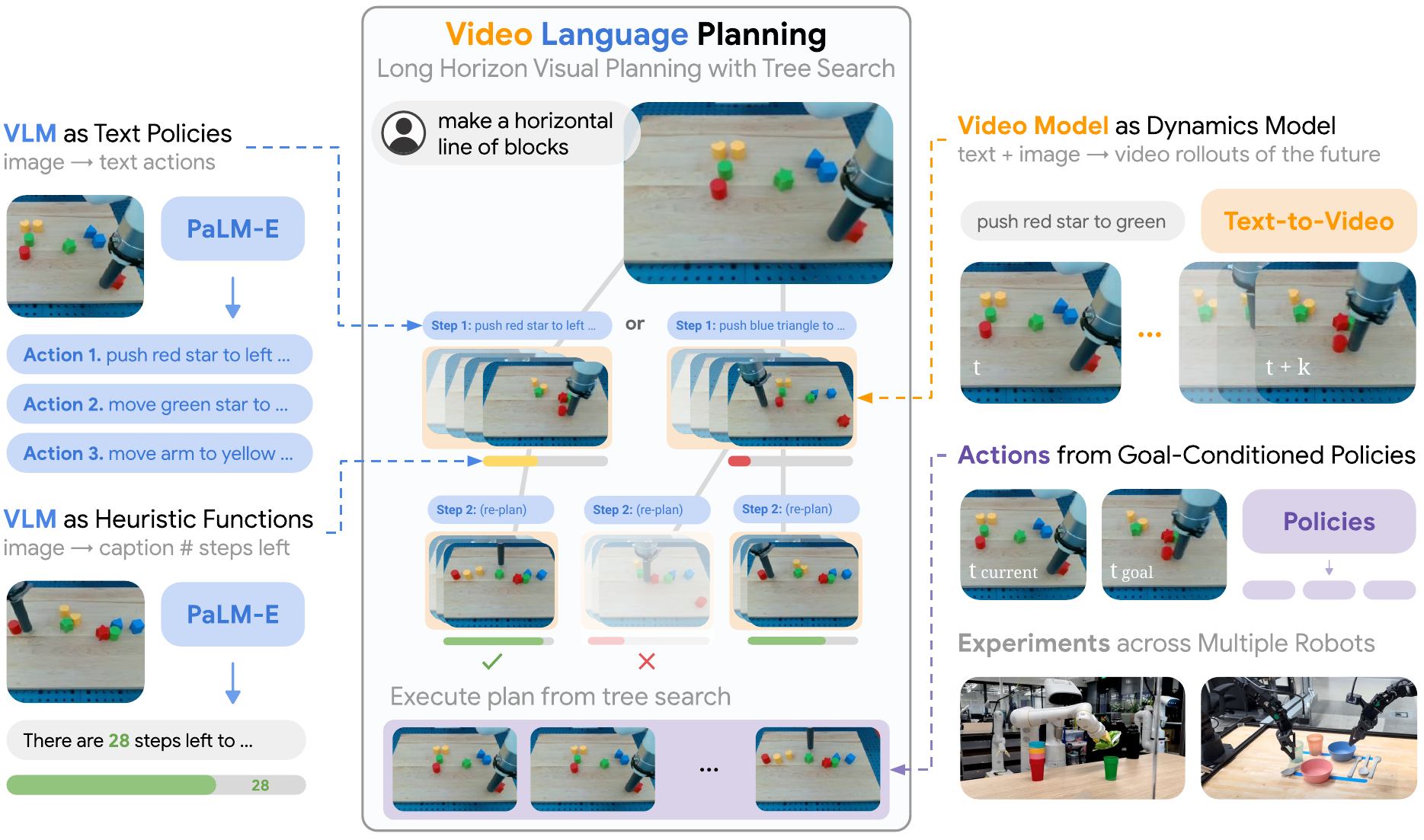}
\end{center}
\vspace{-10pt}
\caption{\textbf{Video Language Planning} uses forward tree search via vision-language models and text-to-video models to construct long-horizon video plans. From an image observation, the VLM policy (top left) generates next-step text actions, which a video model converts into possible future image sequences (top right). Future image states are evaluated using a VLM heuristic function (bottom left), and the best sequence is recursively expanded with tree search (middle). Video plans can be converted to action execution with goal-conditioned policies (bottom right).}
\vspace{-3pt}
\label{fig:teaser}
\end{figure}

In this work, we propose to integrate vision-language models and text-to-video models to enable \textit{video language planning} (VLP), where given the current image observation and a language instruction, the agent uses a VLM to infer high-level text actions, and a video model to predict the low-level outcomes of those actions. Specifically, VLP (illustrated in \fig{fig:teaser}) synthesizes video plans for long-horizon tasks by iteratively: (i) prompting the VLM as a policy to generate multiple possible next-step text actions, (ii) using the video model as a dynamics model to simulate multiple possible video rollouts for each action, and (iii) using the VLM again but as a heuristic function to assess the favorability of each rollout in contributing task progress, then recursively re-planning with (i). 

\looseness=-1
The combination of both models enables forward tree search over the space of possible video sequences to discover long-horizon plans (of hundreds of frames) that respect visual dynamics. In particular, VLP offers advantages in that it (i) can generate higher quality plans at inference time by expanding the branching factor of the search, allowing plan quality to scale with increasing compute budget, and (ii) benefits from training on incomplete language-labeled video data, which may contain short-horizon snippets (that can be re-composed and sequenced into long-horizon ones), or segments of videos with missing language labels (but contain dynamics that the video model can still learn from).

Experiments in both simulated and real settings (on 3 robot hardware platforms) show that VLP generates more complete and coherent multimodal plans than baselines, including end-to-end models trained directly to generate long videos conditioned on long-horizon task instructions. VLPs exhibit improved grounding in terms of the consistency of scene dynamics in video plans, and when used in conjunction with inverse dynamics models or goal-conditioned policies to infer control trajectories \citep{du2023learning}, can be deployed on robots to perform multi-step tasks -- from picking and stowing a variety of objects over countertop settings, to pushing groups of blocks that rearrange them into new formations. In terms of execution success, VLP-based systems are far more likely to achieve task completion for long-horizon instructions than state-of-the-art alternatives, including PaLM-E \citep{driess2023palm} and RT-2 \citep{brohan2023rt} directly fine-tuned for long-horizon tasks. We also observe that when co-trained on Internet-scale data, VLP generalizes to new objects and configurations.

Our main contributions are: (i) video language planning, a scalable algorithm for generating long-horizon video plans by synergising vision-language models and text-to-video models, 
(ii) experiments that show how VLP enables both simulated and real robots to perform complex long-horizon manipulation tasks, exhibiting task completion rates that often exceed that of the next best available approach by a significant margin,
and (iii) ablations that study modes of generalization, as well as how VLP scales with more compute.
VLP presents a modern re-examination of visual planning by integrating large generative models pretrained on Internet-scale data, and is not without limitations. We discuss these in \sect{sec:conclusion} and videos will be made available at: \href{https://video-language-planning.github.io/}{https://video-language-planning.github.io/}

\section{Video Language Planning}

Our planning system, \model{}, takes a visual observation $x_0$ of a scene and a natural language goal $g$ and infers a video plan $\{x_t\}_{1:T}$, where each image $x_t$ is as a sub-goal to accomplish $g$.  We assume that each image $x_t$ serves as an accurate representation of world state and use a image goal-conditioned policy as a controller to infer low level control actions $u$ to reach each image state. 

Below, we first discuss how vision-language and video models are used in \model{} as planning submodules \sect{sect:synergy}. Next, we talk about our tree-search algorithm using vision-language and video building blocks in \sect{sect:plan}. Finally, in \sect{sect:action_regression}, we discuss how we can convert video plans into policies to accomplish each long-horizon task.

\subsection{Using Vision-Language and Video Models as Planning Submodules}
\label{sect:synergy}

\looseness=-1
We first discuss how to use vision-language and video models as sub-modules in \model to synthesize long horizon plans. At a high level, we use the multimodal processing power of VLMs to propose abstract 
text actions $a^i$ to execute given goals and images.
We then use the dynamics knowledge of video models to accurately synthesize possible future world states $x_{1:T}^i$ when abstract actions are executed. Finally, we use a VLM to process possible future world states $x_{1:T}^i$ and assess which sequence $x_{1:T}$ and associated actions are the most promising to complete a task. 

\looseness=-1
\myparagraph{Vision-Language Models as Policies.}  Given a high-level goal $g$, \model searches over a space of possible abstract actions;  these text actions $a$ are generated by a VLM policy $\pi_{\text{VLM}}(x, g) \rightarrow a$ that is conditioned both on the goal $g$ and an image $x$ of the current state.  We implement this policy following ~\citet{driess2023palm} and query the VLM for a natural language action to take given as context the natural language goal and a tokenized embedding of the current image (\fig{fig:teaser} Top Left).  We experiment with two different strategies for constructing this policy. In the first, we provide the VLM a set of example text action labels and ask the VLM to predict possible actions to accomplish a goal.  In the second, we finetune the PaLM-E model on randomly selected short trajectory snippets $x_{1:S}$ labeled with abstract actions 
inside a long trajectory $x_{1:H}$ that accomplishes a long horizon goal $g$.

\myparagraph{Video Models as Dynamics Models.} In order to perform the high-level search, given an image $x$ of a current state and a language description of an abstract action, we need to predict the concrete resulting state.  In addition, to generate low-level controls that instantiate this abstract action, we need a feasible sequence of low-level states that ``interpolate'' between the current state and the resulting state.  We obtain both of these things from a text-to-video model $f_{\text{VM}}(x, a)$, which takes an image $x$ and a short horizon text instruction $a$ and outputs a short synthesized video $x_{1:S}$ starting at the image observation $x_0$ (\fig{fig:teaser} Top Right) following  ~\citet{du2023learning}. We construct this text-to-video model by training on a set of short image trajectory snippets $x_{1:T}$ and associated language labels $a$. 

\myparagraph{Vision-Language Models as Heuristic Functions.} To effectively prune branches in search, we use a VLM to implement a heuristic function $H_{\text{VLM}}(x, g)$ which takes as input an image observation $x$ and a natural language goal description $g$ and outputs a scalar ``heuristic'' predicting the number of actions required to reach a state satisfying goal $g$ from current state $x$ (\fig{fig:teaser} Bottom Left). To construct this heuristic function, we finetune a PaLM-E model using long trajectory snippets $x_{1:H}$ which accomplish a long horizon goal $g$, and train it to predict, given an image in the subtrajectory $x_t$, the number of steps left until the end of the trajectory snippet. The negated number of predicted steps to goal completion from VLM is used to implement $H_{\text{VLM}}(x, g)$ (so that high heuristic value corresponds to being close to goal completion).

\begin{figure}[t]
\centering
\begin{minipage}{\linewidth}
\begin{algorithm}[H]
    \begin{algorithmic}[1]
    \STATE \small{\textbf{Input:} Current visual observation $x_0$, Language goal $g$}
    \STATE \small{\textbf{Functions:} VLM Policy $\pi_{\text{VLM}}(x, g)$, Video Model $f_{\text{VM}}(x, a)$, VLM Heuristic Function $H_{\text{VLM}}(x, g)$} 
    \STATE \small{\textbf{Hyperparameters:} Text-Branching factor $A$, Video-Branching factor $D$, Planning Beams $B$, Planning horizon $H$} \\
    \STATE plans $\leftarrow [ \hspace{0.1em} [x_0] \hspace{0.5em} \forall \hspace{0.5em} i \in \{1 \ldots B\}]$ \hspace{4.50cm} \small{\color{gray}{\# Initialize B Different Plan Beams}}
    \FOR{$h = 1 \ldots H$}
        \FOR{$b = 1 \ldots B$}
            \STATE $x \gets \text{plans}[b][-1]$ \hspace{4.25cm} \small{\color{gray}{\# Get the Latest Image State in the Plan Beam}}
            \STATE $a_{1:A} \gets \pi(x, g)$ \hspace{5.75cm} \small{\color{gray}{\# Generate $A$ Different Text Actions}}
            \STATE video\_branches $\gets [f_{\text{VM}}(a, a_i)$ for i in $(1 \ldots A)$ for j in $(1 \ldots D)]$
            \STATE plans[b].append(argmax(video\_branches, $H_{\text{VLM}}$)) \hspace{1cm} \small{\color{gray}{\# Add Video with Highest Value to Plan}}
        \ENDFOR
        \STATE max\_idx, min\_idx $\gets$ argmax(plans, $H_{\text{VLM}}$),  argmin(plans, $H_{\text{VLM}}$)
        \STATE plans[min\_idx] $\gets$ plans[max\_idx] \hspace{2.55cm} \small{\color{gray}{\# Periodically Replace the Lowest Value Plan}}
    \ENDFOR
    \STATE plan $\gets$ argmax(plans, $H_{\text{VLM}}$) \hspace{5.6cm} \small{\color{gray}{\# Return Highest  Value Plan}}
    \end{algorithmic}
    \caption{\small Decision Making with \model{}}
    \label{alg:cond_gen}
\end{algorithm}
\end{minipage}
\end{figure}
\subsection{Planning with Vision-Language Models and Video Models}
\label{sect:plan}

Given a combination of modules discussed in \sect{sect:synergy}, directly applying the $\pi_{\text{VLM}}$ to infer text actions $a$ to reach goal $g$ is not sufficient, as $\pi_{\text{VLM}}$ is not able to perform sufficiently accurate long-horizon reasoning to select actions that are helpful in the long run. Furthermore, there are many possible low-level image sub-sequences that correspond to different ways to perform $a$, but it is critical to select one that is consistent with the rest of the actions that must be taken to reach $g$.

Instead, we propose to search for a sequence of actions to reach $g$, corresponding to finding a long-horizon video plan $x_{1:H}$ which optimizes 
\begin{equation}
 x_{1:H}^{*} = \argmax_{x_{1:H} \sim f_{\text{VM}}, \pi_{\text{VLM}}} H_{\text{VLM}}(x_H, g).   
 \label{eqn:optimize}
\end{equation}
A separate procedure is then used to instantiate control actions $u$ to enact the optimized video plan $x_{1:H}^{*}$.  To sample long-horizon video plans $x_{1:H}$, we first synthesize a short horizon video plan $x_{1:S}$ from a starting image $x$ through $x_{1:S} = f_{\text{VM}}(x, \pi_{\text{VLM}}(x, g))$ and autoregressively extend to a full long-horizon video plan by recursively applying $f_{\text{VM}}(x, \pi_{\text{VLM}}(x, g))$ on the final synthesized image state $x_S$. To optimize across video plans in \eqn{eqn:optimize}, we use a tree-search procedure based on parallel hill climbing~\citep{selman2006hill} (illustrated in Algorithm~\ref{alg:cond_gen}). 

Our planning algorithm initializes a set of $B$ parallel video plan beams. At each step of the planning horizon, for each video beam, we first sample a set of $A$ actions using $\pi_{\text{VLM}}(x, g)$, and for each action we synthesize $D$ different videos using $f_{\text{VM}}(x, a)$. We then use our heuristic function $H_{\text{VLM}}(x, g)$ to select the generated video with the highest heuristic among the $A \times D$ generated videos and extend the corresponding video plan beam with this generated video. Over the course of plan generation, certain video plan beams will  obtain high heuristic value and be more promising to explore. Therefore, every 5 steps, we discard the beam with the lowest value and replicate its video plan with the beam with the highest value. Our final full long horizon video plan corresponds to the beam with highest heuristic value at the end of planning. 

\myparagraph{Preventing Exploitative Model Dynamics.} When our planning procedure optimizes the VLM heuristic function $H_{\text{VLM}}(x, g)$ it can exploit irregularities in the dynamics model $f_{\text{VM}}(x, a)$ to get artificially high estimates. For instance, the planning procedure can exploit videos from $f_{\text{VM}}(x, a)$ where key objects have teleported to desired locations or where the final image observation obscures undesirable portions of world state. To prevent over-exploitation of $H_{\text{VLM}}(x,g)$, during the planning procedure in Algorithm~\ref{alg:cond_gen}, we discard generated videos from $f_{\text{VM}}(x, a)$ if they increase the the heuristic estimate $H_{\text{VLM}}(x, g)$ above a fixed threshold.

\begin{figure}[t]
    \centering
    \includegraphics[width=\linewidth]{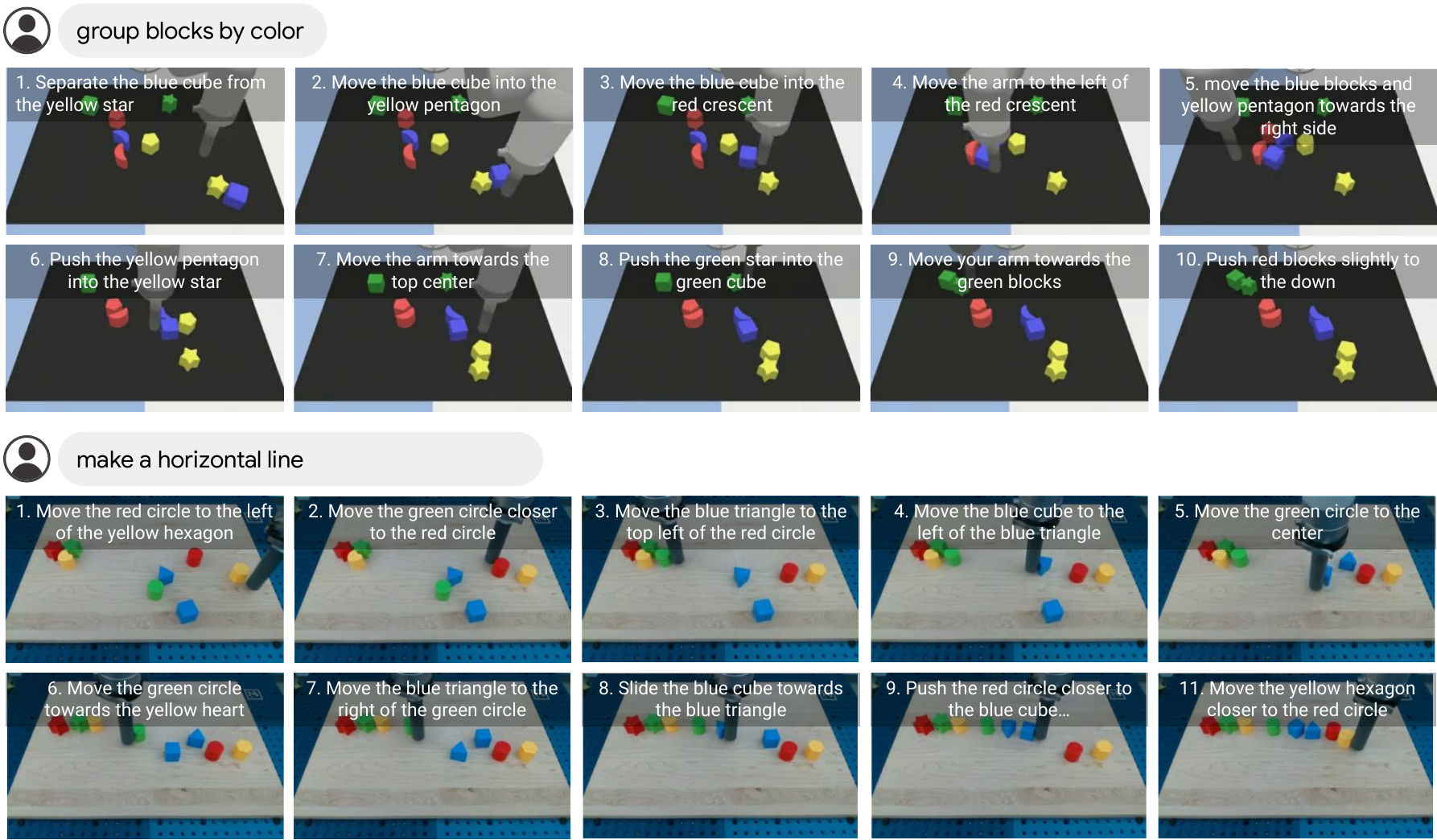}
    \vspace{-15pt}
\caption{\textbf{Long Horizon Video Plan.} Long horizon video plans generated by \model on both simulated and real images. \model is {\it only given} the {\it initial image} and  {\it language goal}. Language subplans and other image frames are {\it directly synthesized}.}
\label{fig:video_gen_blocks}
\vspace{-5pt}
\end{figure}

\subsection{Action Regression From Video through Goal-Conditioned Policies}
\label{sect:action_regression}

Given a synthesized video plan $x_{1:H}$, to execute tasks, we must infer control actions $u$ to reach synthesized images. Prior work infers actions from synthesized videos by using an inverse-dynamics model on each synthesized frame~\citep{du2023learning}.

In many settings, a single action may not be sufficient to directly reach the next synthesized image, {\it i.e.} if you need to remove the cap off a toothbrush, and even in settings in which this is the case, it may be difficult to precisely predict the correct action to reach the next frame. To reduce the burden on the inverse dynamics model, we propose to use a short-horizon goal-conditioned policy $\pi_{\text{control}}(x, x_g)$, which  given the current image observation $x$ and next frame in a video plan $x_g$ outputs a low level control action $u$ that makes progress towards $x_g$. For each frame in our video plan $x_{1:H}$, the goal-conditioned policy is executed for a fixed pre-specified number of timesteps. We train $\pi_{\text{control}}$ using paired image and low level control snippets $x_{1:T}^i$ and $u_{1:T}^i$, where we sample a random timestep $t$, a corresponding state $x_t$, and future state $x_{t+h}$, and train $\pi_{\text{control}}(x_t, x_{t+h})$ to predict $u_t$.

\myparagraph{Replanning.} Given a very long-horizon task, it is both difficult to use $\pi_{\text{control}}$ to accurately execute the full video plan $x_{1:H}$ (due to accumulating error) and difficult to fully synthesize a plan that completely finishes a long-horizon task given a fixed planning horizon.  To circumvent this issue, we use receding horizon control 
 strategy~\citep{kwon2005receding}, where we  generate videos plans with a fixed horizon (that might not fully complete the task), and then repeatedly regenerate/replan video plans with the same horizon after a fixed number of action executions.

\begin{table*}[t]
\small\setlength{\tabcolsep}{5.5pt}
\centering
\begin{tabular}{l|ccc|ccc}
      & \multicolumn{3}{c}{\bf Sim Environment} &  \multicolumn{3}{c}{\bf Real Environment} \\
      \cmidrule(lr){2-4} \cmidrule(lr){5-7} 
       & Move  & Group  & Make & Move & Group & Make  \\
      Model & Area &  Color & Line & Area & Color & Line \\
      \midrule
      UniPi & 2\%  & 4\% & 2\%  &  4\% & 12\% & 4\% \\
      \model (No Value Function)  &  10\% & 42 \%  & 8\% &  20\% & 64\% & 4\% \\
      \model (Ours)  & {\bf 58\%}  & {\bf 98\%}  & {\bf 66\%}  &  {\bf 78\%}  & {\bf 100\%} & {\bf 56\%} \\
    \bottomrule
\end{tabular}
\vspace{-5pt}
\caption{\small \textbf{Accuracy of Generated Video Plans.} The percentage \model and baselines are able to synthesize a full video plan which can fully complete tasks in simulation and real environments. \model substantially outperforms both UniPi and directly combining the VLM policy }
\label{tbl:video_accuracy}
\end{table*}

\section{Experimental Results}

\begin{figure*}[t]
\setlength{\tabcolsep}{5.5pt}

\begin{minipage}{0.5\textwidth}
\centering
\scalebox{0.9}{
\begin{tabular}{ccc|c}
       & Language &  Video &  Line \\
      Beams & Branch &  Branch & Performance \\
      \midrule
      1 &  1 & 1 &  4\% \\
      1 &  1 & 4 &  10\% \\
      1 &  4 & 4 & 22\%   \\
      2 &  4 & 4 &  {\bf 56\%} \\
    \bottomrule
\end{tabular}
}
\end{minipage}
\hfill
\begin{minipage}{0.5\textwidth} %
    \centering
    \includegraphics[width=\textwidth]{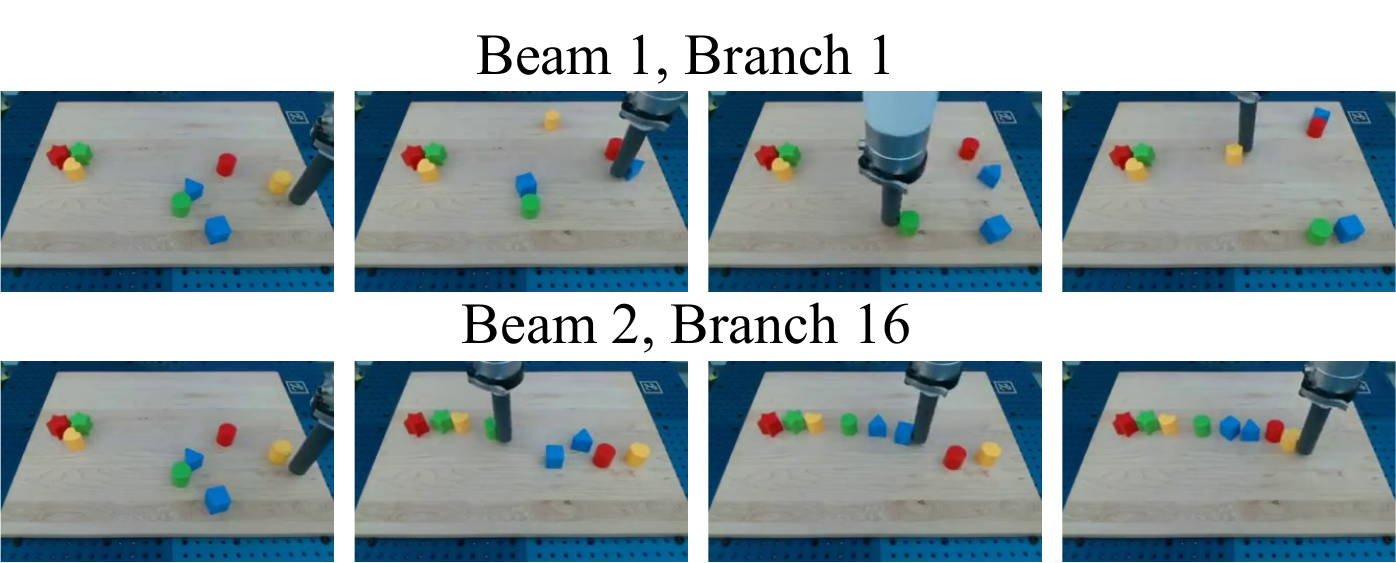} 
\end{minipage}
\vspace{-5pt}
\caption{ \looseness=-1 \small \textbf{Video Accuracy vs Planning Budget.} \textit{Left}: \model scales positively with more compute budget; it is better able to synthesize plans to solve tasks with more planning (i.e. with a higher beam-search branching factor). Success percentage reported on the make line task. \textit{Right}: Qualitative illustration of video plans for making a line generated without planning (Beam 1, Branch 1) compared to extensive planning (Beam 2, Branch 16). }
\label{tbl:plan_video_ablation}
\end{figure*}

\begin{figure}[t]
    \centering
    \includegraphics[width=\linewidth]{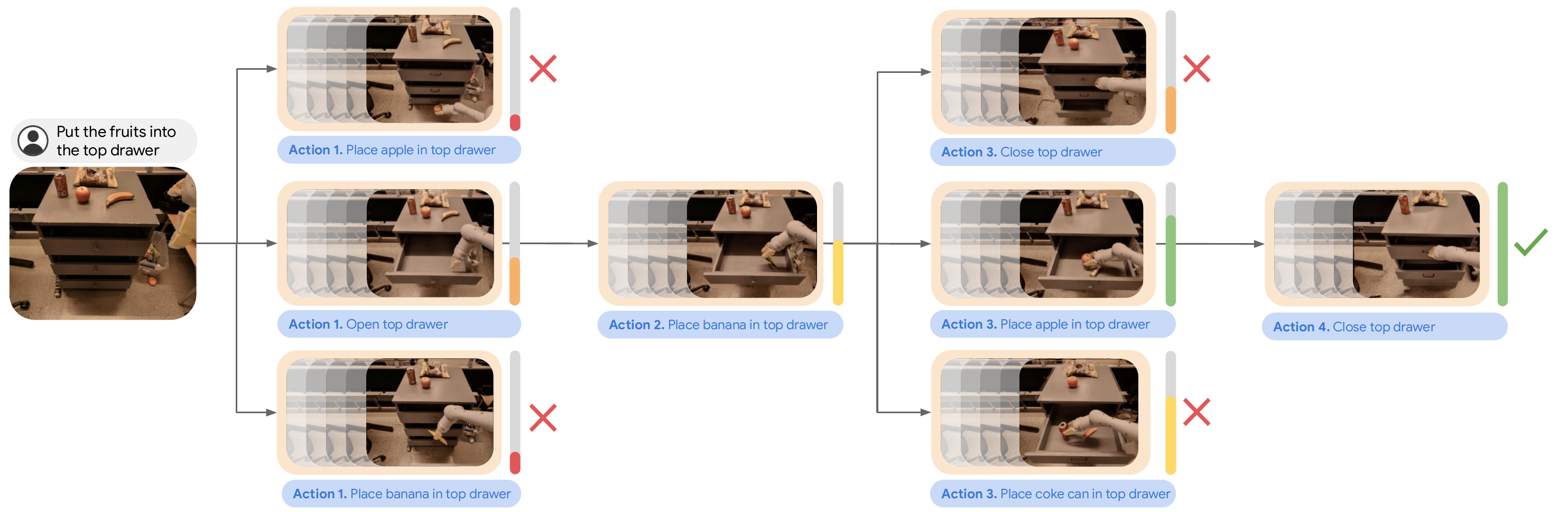}
    \vspace{-15pt}
\caption{\textbf{Planning Tree on 7DoF Mobile Manipulator.}  \model is able to prune unlikely language and video branches to synthesize a coherent long-horizon video plan.}
\label{fig:video_fractal}
\end{figure}
\begin{figure}[t]
    \centering
    \includegraphics[width=\linewidth]{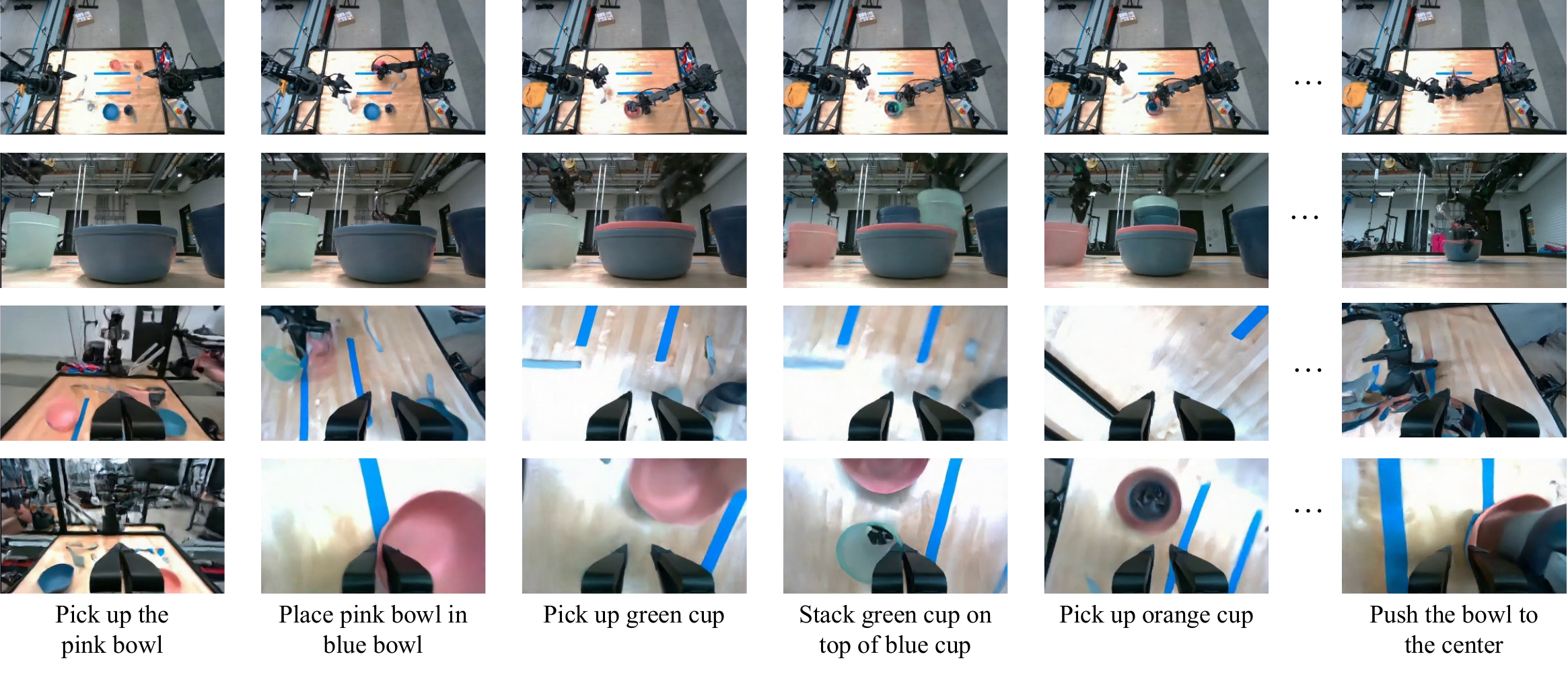}
    \vspace{-20pt}
\caption{\textbf{Multiview Video Plans for Dexterous Manipulation.} Long horizon video plans (and associated language subgoals) generated by \model for solving the long horizon task of stacking everything in a table together. \model is able to synthesize multiview video plans across 4 cameras, that are consistent with each other and with task completion. The first 5 generated language subgoals goals are illustrated as well as the final generated goal image. \model is only given first image.}
\label{fig:video_aloha}
\vspace{-5pt}
\end{figure}

We first evaluate the ability of \model to synthesize long-horizon video plans for different tasks in \sect{sect:video_plan}. We then investigate \model's ability to %
execute generated video plans in various environments in \sect{sect:action}. Finally, we further investigate generalization capabilities of \model in \sect{sect:generalization}.

\subsection{Long-Horizon Video Synthesis}

\begin{table*}[t]
\small\setlength{\tabcolsep}{5.5pt}
\centering
\begin{tabular}{lcccccc}
      & \multicolumn{2}{c}{\bf Move to Area} &  \multicolumn{2}{c}{\bf Group by Color} & \multicolumn{2}{c}{\bf Make Line} \\
      \cmidrule(lr){2-3} \cmidrule(lr){4-5} \cmidrule(lr){6-7}
      Model & Reward & Completion & Reward & Completion & Reward & Completion\\
      \midrule
      UniPi ~\citep{du2023learning} & 30.8 &  0\% & 44.0  & 4\% & 44.0 & 4\% \\
      LAVA ~\citep{lynch2023interactive}  & 59.8 & 22\% & 50.0 & 2\% & 33.5  & 0\%  \\
      RT-2  ~\citep{brohan2023rt} & 18.5 & 0\% & 46.0  & 26\% & 36.5 & 2\% \\
      PALM-E ~\citep{driess2023palm} &  36.5 &  0\% & 43.5 & 2\% & 26.2 & 0\%  \\
      \model (Ours) & {\bf 87.3} & {\bf 64\%} & {\bf 95.8} & {\bf 92\%} & \bf{65.0} & \bf{16\%}  \\
    \bottomrule
\end{tabular}
\vspace{-5pt}
\caption{\small \textbf{Execution Performance on Long Horizon Tasks.} \model is able to accurately execute actions for different long-horizon synthetic language table tasks. \model substantially outperforms all existing methods.}
\label{tbl:language_table_sim}
\end{table*}
\begin{figure}[t!]
    \centering
    \includegraphics[width=\linewidth]{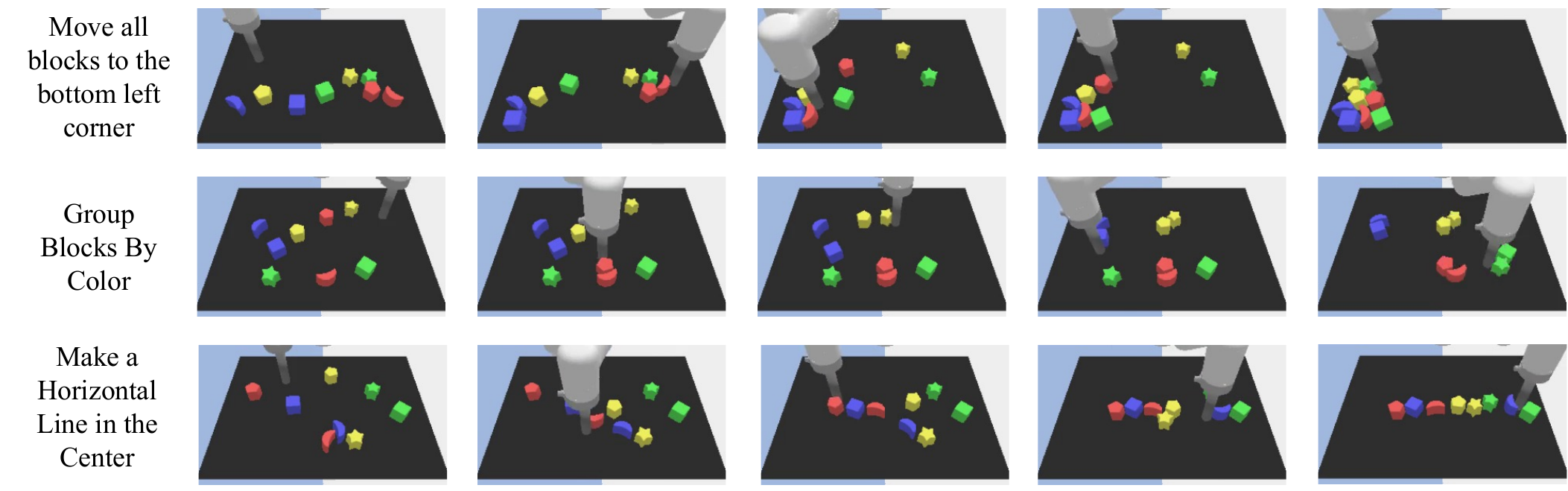}
    \vspace{-15pt}
\caption{\textbf{Simulation Execution.} Illustration of execution of \model on different simulated environments. \model is able to accomplish different long horizon goals.}
\label{fig:sim_execution}
\end{figure}

\label{sect:video_plan}

\myparagraph{Baselines.} We compare our approach with two other approaches for synthesizing long-horizon video plans. First, we consider training a text-to-video model $f_{\text{VM}}$ on %
long horizon text goals, as in UniPi ~\citep{du2023learning}, omitting the entire VLP planning process. Next, we consider synthesizing long horizon video plans by chaining $\pi_{\text{VLM}}$ policy with $f_{\text{VM}}$, without the heuristic function.

\myparagraph{Object Rearrangement.} We first illustrate video plans in the Language Table environment~\citep{lynch2023interactive}. We give as input to \model a random image and randomly chosen language goal. We then visualize the generated VLP plans (\fig{fig:video_gen_blocks}).  We report  the quantitative success of synthesizing long-horizon videos given random starting images for each task in Language Table in \tbl{tbl:video_accuracy}. For each reported number, we generated a total of 50 videos from each method and visually assessed the percentage of time the video successfully solved the given task. \model substantially outperforms the baseline of directly synthesizing videos given a long-horizon prompt, indicating the importance of hierarchical structure. \model further outperforms the ablation of only using a VLM policy with a video model, pointing to the effectiveness of the \model planning procedure and including the value function.

\myparagraph{Effect of Search of Video Synthesis.} We analyze the effect of search in generating long-horizon videos in \fig{tbl:plan_video_ablation} (left). We consider increasing the video branching, language branching and the beams in the search procedure. We find that each increase of branching factor in search substantially increases the success of synthesized long horizon plans. A qualitative illustration of the difference of generated plans with small and large branching factor is illustrated in \fig{tbl:plan_video_ablation} (right).

\myparagraph{Planning on 7DoF Mobile Manipulators.} We qualitatively illustrate how we can generate plans on a higher-DoF, 7DoF Mobile Manipulator in \fig{fig:video_fractal}. Our planning system is able to generate videos of actions that both open and close drawers in order to satisfy specified text prompts.

\myparagraph{Planning on Multicamera 14DoF Bi-Manual Manipulators.} We further illustrate how our approach can generate multi-view 4-camera videos of dexterous manipulation on the 14DoF bi-manual ALOHA~\citep{zhao2023learning} platform in \fig{fig:video_aloha}. Our video model outputs videos across views simultaneously (by concatenating each view channelwise), while our VLM policy and heuristic function takes as input top and side views.  Our approach is able to synthesize multiview consistent plans which are able to both stack bowls, cups, and utensils.

\begin{figure}[t]
    \centering
    \includegraphics[width=\linewidth]{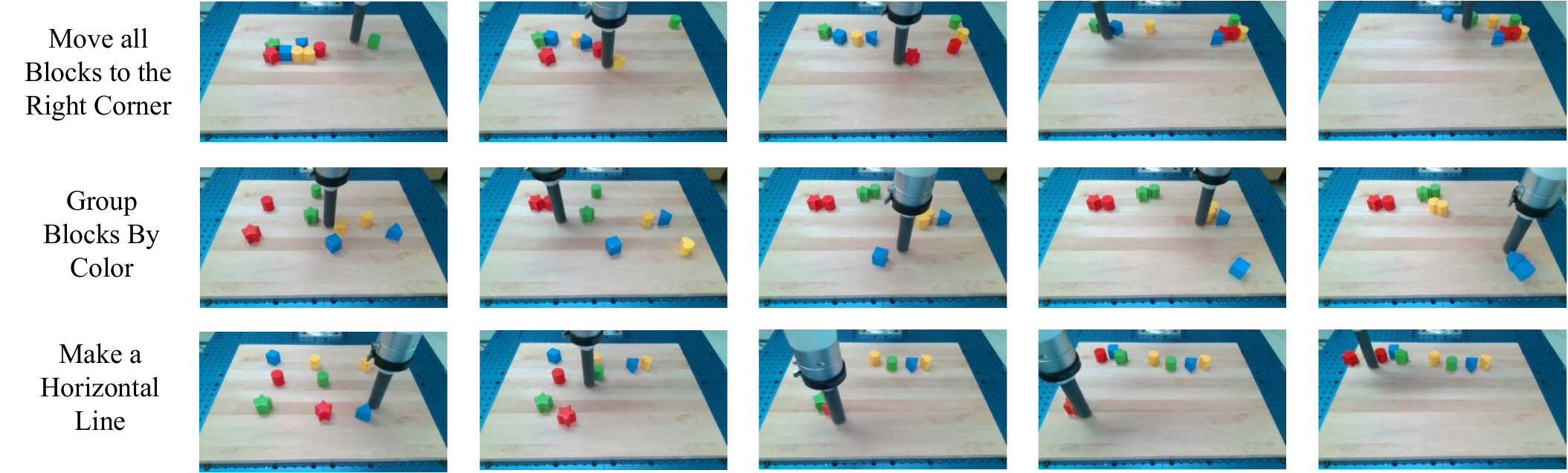}
    \vspace{-15pt}
\caption{\textbf{Real Execution.} Illustration of execution \model on real world robots. \model is able to accomplish different long horizon goals when executed on real robots.}
\label{fig:real_execution}
\end{figure}
\begin{figure}[t]
    \centering
    \includegraphics[width=\linewidth]{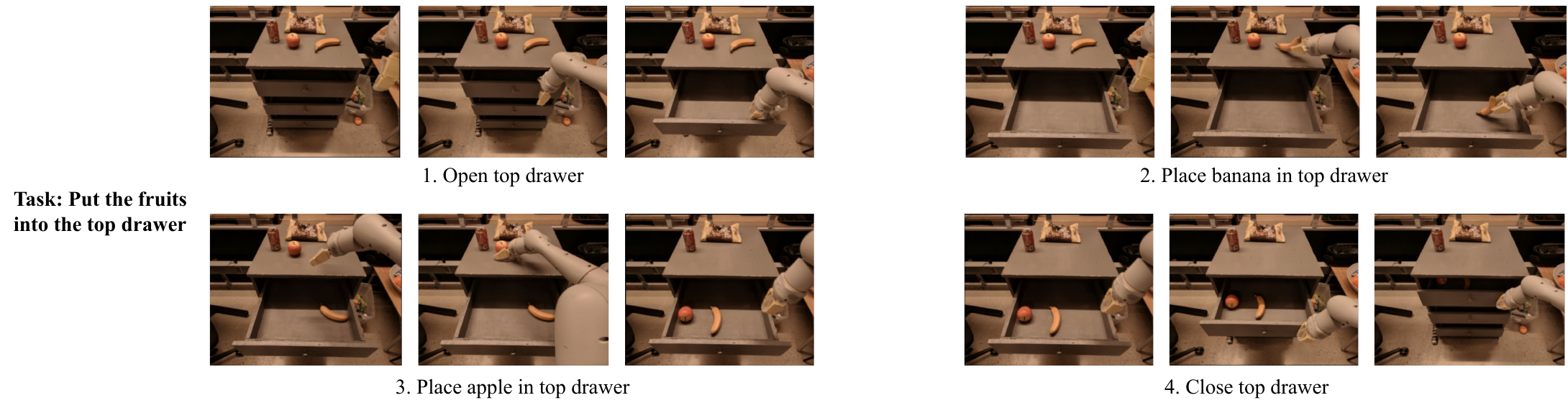}
    \vspace{-20pt}
\caption{\textbf{7DoF Mobile Robot Execution.} \model is able to execute complex, long horizon plans on mobile robot. }
\label{fig:fractal_execution}
\end{figure}

\subsection{Long-Horizon Execution}
\label{sect:action}

We next evaluate the ability of \model to not only generate plans as in \sect{sect:video_plan}, but to actually use  planning (and replanning) to {\em{execute}} long-horizon tasks in closed-loop environments.

\myparagraph{Baselines.} We compare our approach to a set of approaches to solve long-horizon tasks. (i) We consider using a VLM to directly plan, using PaLM-E ~\citep{driess2023palm} to plan short horizon text snippets to execute, which are converted to actions using a text-conditioned policy, conditioned on generated text snippets from PaLM-E. We also (ii) compare with UniPi ~\citep{du2023learning}, where videos are directly generated by a text-to-video model trained on long-horizon text goals and converted to actions using our goal-conditioned policy. Next, we consider (iii) directly learning language-conditioned behavioral cloning policy on long-horizon text and actions, using the codebase and architecture of the LAVA model ~\cite{lynch2023interactive}. Finally, we (iv) compare with leveraging existing vision-language models for control, and train the RT2 model ~\cite{brohan2023rt} on long-horizon text and actions. 

\myparagraph{Quantitative Results.} We evaluate each approach quantitatively on moving all blocks to different areas of the board, grouping blocks by color, or making blocks in a line (details on quantitative evaluation in Appendix).  We report quantitative results in \tbl{tbl:language_table_sim}, and find that our approach substantially outperforms all baseline methods. As the task horizon of each task is very long (around 1500 steps), we found that many baseline methods would become ``stuck" and stop acting effectively. We illustrate example executions using \model{} in \fig{fig:sim_execution}.

\myparagraph{Effect of Planning.} Next, we analyze the effect of the amount of planning on execution success rates in \tbl{tbl:execution_ablation}. We find that increasing both the planning horizon and the branching factor of planning substantially improves the success of task execution (at the cost of inference time).

\myparagraph{Ablations of Goal-Conditioned Policy} We further conduct experiments on different approaches to extracting actions from videos in \tbl{tbl:action_ablation}. We find that using a goal-conditioned policy conditioned on each intermediate frame in a synthesized video leads to the best overall performance (outperforming using a goal-conditioned policy sparsely on the end frames of each short-horizon video).

\begin{table}[t!]
\begin{minipage}[b]{0.48\linewidth}
\centering
\small\setlength{\tabcolsep}{5.5pt}
\scalebox{0.9}{
\begin{tabular}{ccc|cc}
       & Planning &  Branching &  Line & Line\\
      Beams & Horizon &  Factor & Score & Completion\\
      \midrule
      1 &  1 & 4 & 48.9 & 0\%   \\
      1 &  1 & 16 & 53.3  & 2\% \\
      1 &  2 & 16 &  58.1 & 8\% \\
      2 &  2 & 16 & {\bf 65.0} & 16\% \\
    \bottomrule
\end{tabular}
}
\caption{\small \textbf{Execution Accuracy vs Planning Budget.} \model is able to more accurately execute video plans to solve tasks with a larger amount of planning. Success percentage reported on the make line task.}
\label{tbl:execution_ablation}
\end{minipage}%
\hfill
\begin{minipage}[b]{0.48\linewidth}
\centering
\small\setlength{\tabcolsep}{5.5pt}
\scalebox{0.95}{
\begin{tabular}{l|cc}
        &  Group Color & Group Color\\
      Action Inference & Score & Completion\\
      \midrule
      Inverse Dynamics & 89.7 & 80\% \\
      Goal Policy (Last) &  85.0 &  66\%\\
      Goal Policy (Every) & {\bf 95.8} & {\bf 92\%} \\
    \bottomrule
\end{tabular}
}
\caption{\small \textbf{Extracting Actions From \model Video Plans.}  Comparison of using inverse dynamics or applying a goal-conditioned policy to either the last frame or every frame of synthesized  short-horizon video. Success percentage reported on the group by color task.}
\label{tbl:action_ablation}
\end{minipage}
\end{table}

\myparagraph{Real Execution.} We provide executions of \model on multiple real-world robots in \fig{fig:real_execution} and \fig{fig:fractal_execution}.  As in \fig{fig:real_execution}, \model is able to effectively execute each shown long-horizon task on a real Language Table robot. We further provide executions of generated video plans of our approach on the 7DoF mobile manipulator in \fig{fig:fractal_execution}. Similarly we find that a goal-conditioned policy can realize plans.

\begin{figure}[t]
    \centering
    \includegraphics[width=\linewidth]{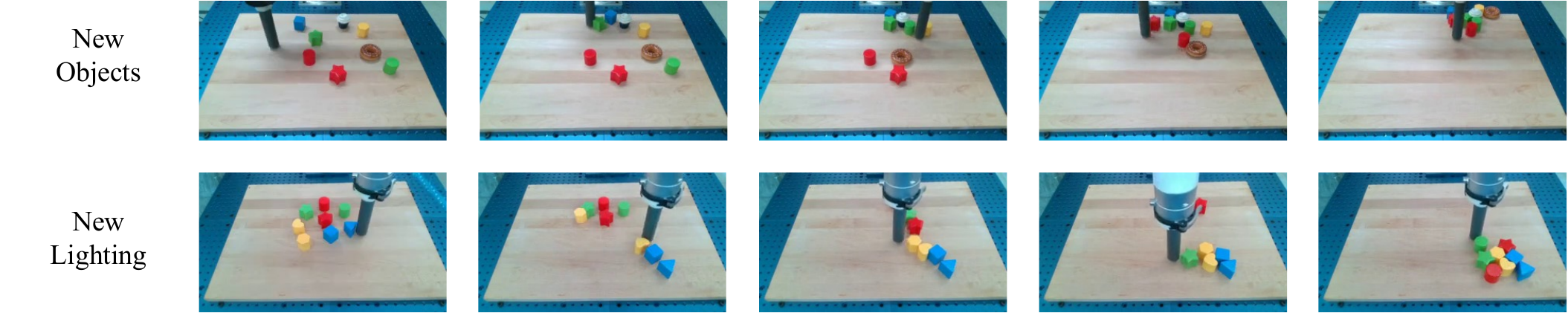}
    \vspace{-15pt}
\caption{\textbf{Generalization to Objects and Lighting.} \model is able to generalize execution to scenes with three new objects (top) consisting of a wooden yellow hexagon, rubber donut and rubber cupcake. \model is able to also generalize to a robot placed in a new office (bottom) with different lighting conditions (a lot more lighting on the right side of the board) and similarly execute tasks. }
\label{fig:generalization}
\end{figure}
\begin{figure}[t]
    \centering
    \includegraphics[width=\linewidth]{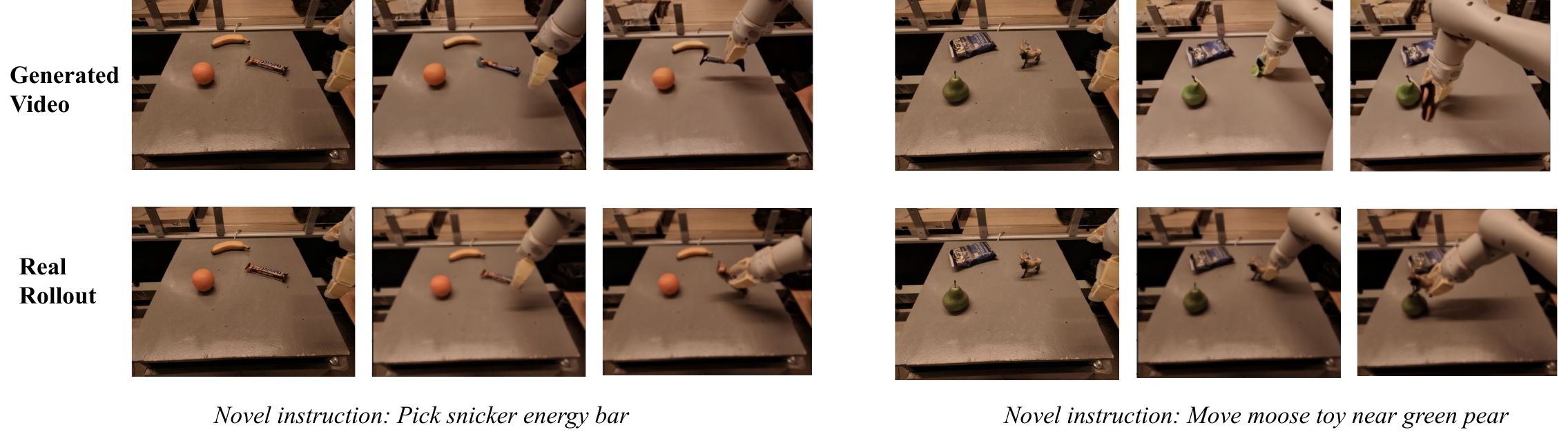}
    \vspace{-20pt}
\caption{\textbf{Task Generalization} \model can generalize to new tasks on unseen objects using internet knowledge. }
\label{fig:task_generalization}
\end{figure}

\subsection{Generalization}
\label{sect:generalization}

\myparagraph{Generalization to Lighting and Objects.} In \model, policy execution is abstracted into visual goal generation followed by a goal-conditioned controller. With this abstraction, a video model can simply focus on capturing the visual dynamics objects, while a goal-conditioned policy needs to focus only on relevant visual details to achieve the next (nearby) goal. We found that this enables \model to generalize well, as the video model is able to visually generalize to new images, while the policy is able to generalize well to nearby new visual goals.
In \fig{fig:generalization} (top), \model is able to generalize the task of putting all objects in top right corner, to three new objects, a rubber donut and cupcake and a wooden hexagon.  In \fig{fig:generalization} (bottom) \model is further able to generalize to lighting conditions substantially different than the ones the model was trained on and can be successfully deployed to a new robot in a different building location.

\myparagraph{Generalization to New Tasks} In \model, both VLM and text-to-video models may be pre-trained on a vast amount of Internet data. In \fig{fig:task_generalization}, we  train both VLM and text-to-video models on a large mix of datasets and illustrate how it further generalizes and executes new tasks on unseen objects.

\section{Related Work}

There is a great deal of recent work in leveraging large pretrained foundation models for decision-making agents~\citep{yang2023foundation} -- in particular, using LLMs (and the commonsense knowledge they store) to infer actions represented in text.
For example, given language instructions, LLMs can be prompted to generate high-level step-by-step plans~\citep{huang2022language,ahn2022can,wang2023describe}
-- each step mapped robot actions, invoked with pre-trained policies or existing APIs~\citep{liang2023code}.
Using existing VLMs~\citep{chen2022pali}, plans can also be conditioned on visual inputs, represented with language descriptions ~\citep{zeng2022socratic,huang2022inner} or token embeddings co-trained with LLMs~\citep{driess2023palm}.
Nevertheless, VLMs are often pretrained on static image datasets (\ie image-in, text-out), and subsequently (i) are limited to plans that can be expressed in text, and (ii) may struggle to reason about the dynamics of the world.

Video models, on the other hand, exhibit potential to generate informative image sequences of the future, and there exists a body of work using them as dynamics models to capture object motion across image states ~\citep{finn2016unsupervised,xue2016visual,oprea2020review,oh2015action}. 
Generating high-quality videos over long time horizons is a known challenge~\citep{babaeizadeh2021fitvid,saxena2021clockwork};
however, recent progress in larger models and text-conditioning give rise to a new generation of text-to-video models that can generate detailed image frames \citep{ho2022imagen,villegas2022phenaki}, that can drive policies for decision making~\citep{du2023learning}. %
However, these works focus predominantly on short-horizon videos.
Our work investigates using text-to-video models together with VLMs to produce long-horizon video plans (potentially predicting hundreds of frames into the future), leveraging the scalability of tree search to plan in the space of videos and language.

Our work can be viewed as bringing together two families of generative foundation models to compose new capabilities -- a strategy shown to be effective for applications such as 2D image synthesis~\citep{du2020compositional,liu2021learning,lace, liu2022compositional,wu2022zeroc,du2023reduce,wang2023concept}, 3D synthesis~\citep{po2023compositional}, video synthesis~\citep{yang2023probabilistic}, trajectory planning~\citep{Du2019ModelBP,urain2021composable,gkanatsios2023energy,yang2023compositional} and multimodal perception~\citep{li2022pre,zeng2022socratic}. 
Most similar to our work, HiP~\citep{ajay2023compositional} combines language models, video models, and action models for hierarchical planning. In contrast, our work uses a forward search procedure to combine the strengths of a VLM and a video model. This composition through forward search enables us to simulate and reason about long horizons of future actions, while HiP only generates and acts on plans one step at a time.

\section{Limitations and Conclusion}
\label{sec:conclusion}

\myparagraph{Limitations.} Our planning approach leverages images as a world state representation. In many tasks, this is insufficient as it does not capture the full 3D state and cannot encode latent factors such as physics or mass. 
Limitations can be partly remedied by generating multi-view videos or by using heuristic function with the full video plan as input.
In addition, we observed that our video dynamics model does not always simulate dynamics accurately. In several situations, we observed that synthesized videos would make objects spontaneously appear or teleport to new locations. We believe that larger video models, additional training data or explicit reinforcement learning feedback for physics~\citep{black2023training} could help solve these problems.

\looseness=-1
\myparagraph{Conclusions.} We have presented \model, an approach to long-horizon decision making by combining VLMs with text-to-video models. We illustrate the power of test-time composition, in this case planning, to generate behaviors much more complex than its components. While our experiments largely explore \model in the context of robotics, there may be downstream applications in other areas too, such as steerable media generation. We believe that future exploration in test-time compositions of  different foundation models can be a fruitful source of gains to construct more complex intelligent systems. 

\myparagraph{Acknowledgements.} We would like to thank Tomas Lozano-Perez for providing helpful comments in the project and Kamyar Ghasemipour for help in setting up experiments on the Language Table real robot. We gratefully acknowledge support from NSF grant 2214177; from AFOSR grant FA9550-22-1-0249; from ONR MURI grant N00014-22-1-2740; from ARO grant W911NF-23-1-0034; from the MIT-IBM Watson Lab; from the MIT Quest for Intelligence; and from the Boston Dynamics Artificial Intelligence Institute. Yilun Du is supported by a NSF Graduate Fellowship.

\bibliography{iclr2024_conference}

\begin{thebibliography}{51}
\providecommand{\natexlab}[1]{#1}
\providecommand{\url}[1]{\texttt{#1}}
\expandafter\ifx\csname urlstyle\endcsname\relax
  \providecommand{\doi}[1]{doi: #1}\else
  \providecommand{\doi}{doi: \begingroup \urlstyle{rm}\Url}\fi

\bibitem[Ahn et~al.(2022)Ahn, Brohan, Brown, Chebotar, Cortes, David, Finn, Gopalakrishnan, Hausman, Herzog, et~al.]{ahn2022can}
Michael Ahn, Anthony Brohan, Noah Brown, Yevgen Chebotar, Omar Cortes, Byron David, Chelsea Finn, Keerthana Gopalakrishnan, Karol Hausman, Alex Herzog, et~al.
\newblock Do as {I} can, not as {I} say: Grounding language in robotic affordances.
\newblock \emph{arXiv preprint arXiv:2204.01691}, 2022.
\newblock URL \url{https://arxiv.org/abs/2204.01691}.

\bibitem[Ajay et~al.(2023)Ajay, Han, Du, Li, Gupta, Jaakkola, Tenenbaum, Kaelbling, Srivastava, and Agrawal]{ajay2023compositional}
Anurag Ajay, Seungwook Han, Yilun Du, Shaung Li, Abhi Gupta, Tommi Jaakkola, Josh Tenenbaum, Leslie Kaelbling, Akash Srivastava, and Pulkit Agrawal.
\newblock Compositional foundation models for hierarchical planning.
\newblock \emph{arXiv preprint arXiv:2309.08587}, 2023.

\bibitem[Babaeizadeh et~al.(2021)Babaeizadeh, Saffar, Nair, Levine, Finn, and Erhan]{babaeizadeh2021fitvid}
Mohammad Babaeizadeh, Mohammad~Taghi Saffar, Suraj Nair, Sergey Levine, Chelsea Finn, and Dumitru Erhan.
\newblock Fitvid: Overfitting in pixel-level video prediction.
\newblock \emph{arXiv preprint arXiv:2106.13195}, 2021.

\bibitem[Black et~al.(2023)Black, Janner, Du, Kostrikov, and Levine]{black2023training}
Kevin Black, Michael Janner, Yilun Du, Ilya Kostrikov, and Sergey Levine.
\newblock Training diffusion models with reinforcement learning.
\newblock \emph{arXiv preprint arXiv:2305.13301}, 2023.

\bibitem[Brohan et~al.(2022)Brohan, Brown, Carbajal, Chebotar, Dabis, Finn, Gopalakrishnan, Hausman, Herzog, Hsu, et~al.]{brohan2022rt}
Anthony Brohan, Noah Brown, Justice Carbajal, Yevgen Chebotar, Joseph Dabis, Chelsea Finn, Keerthana Gopalakrishnan, Karol Hausman, Alex Herzog, Jasmine Hsu, et~al.
\newblock Rt-1: Robotics transformer for real-world control at scale.
\newblock \emph{arXiv preprint arXiv:2212.06817}, 2022.

\bibitem[Brohan et~al.(2023)Brohan, Brown, Carbajal, Chebotar, Chen, Choromanski, Ding, Driess, Dubey, Finn, et~al.]{brohan2023rt}
Anthony Brohan, Noah Brown, Justice Carbajal, Yevgen Chebotar, Xi~Chen, Krzysztof Choromanski, Tianli Ding, Danny Driess, Avinava Dubey, Chelsea Finn, et~al.
\newblock Rt-2: Vision-language-action models transfer web knowledge to robotic control.
\newblock \emph{arXiv preprint arXiv:2307.15818}, 2023.

\bibitem[Brown et~al.(2020)Brown, Mann, Ryder, Subbiah, Kaplan, Dhariwal, Neelakantan, Shyam, Sastry, Askell, et~al.]{brown2020language}
Tom Brown, Benjamin Mann, Nick Ryder, Melanie Subbiah, Jared~D Kaplan, Prafulla Dhariwal, Arvind Neelakantan, Pranav Shyam, Girish Sastry, Amanda Askell, et~al.
\newblock Language models are few-shot learners.
\newblock \emph{Advances in neural information processing systems}, 33:\penalty0 1877--1901, 2020.

\bibitem[Cambon et~al.(2009)Cambon, Alami, and Gravot]{cambon2009hybrid}
Stephane Cambon, Rachid Alami, and Fabien Gravot.
\newblock A hybrid approach to intricate motion, manipulation and task planning.
\newblock \emph{The International Journal of Robotics Research}, 28\penalty0 (1):\penalty0 104--126, 2009.

\bibitem[Chen et~al.(2022)Chen, Wang, Changpinyo, Piergiovanni, Padlewski, Salz, Goodman, Grycner, Mustafa, Beyer, et~al.]{chen2022pali}
Xi~Chen, Xiao Wang, Soravit Changpinyo, AJ~Piergiovanni, Piotr Padlewski, Daniel Salz, Sebastian Goodman, Adam Grycner, Basil Mustafa, Lucas Beyer, et~al.
\newblock Pali: A jointly-scaled multilingual language-image model.
\newblock \emph{arXiv preprint arXiv:2209.06794}, 2022.

\bibitem[Chowdhery et~al.(2022)Chowdhery, Narang, Devlin, Bosma, Mishra, Roberts, Barham, Chung, Sutton, Gehrmann, et~al.]{chowdhery2022palm}
Aakanksha Chowdhery, Sharan Narang, Jacob Devlin, Maarten Bosma, Gaurav Mishra, Adam Roberts, Paul Barham, Hyung~Won Chung, Charles Sutton, Sebastian Gehrmann, et~al.
\newblock Palm: Scaling language modeling with pathways.
\newblock \emph{arXiv preprint arXiv:2204.02311}, 2022.

\bibitem[Damen et~al.(2018)Damen, Doughty, Farinella, Fidler, Furnari, Kazakos, Moltisanti, Munro, Perrett, Price, et~al.]{damen2018scaling}
Dima Damen, Hazel Doughty, Giovanni~Maria Farinella, Sanja Fidler, Antonino Furnari, Evangelos Kazakos, Davide Moltisanti, Jonathan Munro, Toby Perrett, Will Price, et~al.
\newblock Scaling egocentric vision: The epic-kitchens dataset.
\newblock In \emph{Proceedings of the European conference on computer vision (ECCV)}, pp.\  720--736, 2018.

\bibitem[Driess et~al.(2023)Driess, Xia, Sajjadi, Lynch, Chowdhery, Ichter, Wahid, Tompson, Vuong, Yu, et~al.]{driess2023palm}
Danny Driess, Fei Xia, Mehdi~SM Sajjadi, Corey Lynch, Aakanksha Chowdhery, Brian Ichter, Ayzaan Wahid, Jonathan Tompson, Quan Vuong, Tianhe Yu, et~al.
\newblock Palm-e: An embodied multimodal language model.
\newblock \emph{arXiv preprint arXiv:2303.03378}, 2023.

\bibitem[Du et~al.(2019)Du, Lin, and Mordatch]{Du2019ModelBP}
Yilun Du, Toru Lin, and Igor Mordatch.
\newblock Model based planning with energy based models.
\newblock \emph{CORL}, 2019.

\bibitem[Du et~al.(2020)Du, Li, and Mordatch]{du2020compositional}
Yilun Du, Shuang Li, and Igor Mordatch.
\newblock Compositional visual generation with energy based models.
\newblock \emph{Advances in Neural Information Processing Systems}, 33:\penalty0 6637--6647, 2020.

\bibitem[Du et~al.(2023{\natexlab{a}})Du, Durkan, Strudel, Tenenbaum, Dieleman, Fergus, Sohl-Dickstein, Doucet, and Grathwohl]{du2023reduce}
Yilun Du, Conor Durkan, Robin Strudel, Joshua~B Tenenbaum, Sander Dieleman, Rob Fergus, Jascha Sohl-Dickstein, Arnaud Doucet, and Will Grathwohl.
\newblock Reduce, reuse, recycle: Compositional generation with energy-based diffusion models and mcmc.
\newblock \emph{arXiv preprint arXiv:2302.11552}, 2023{\natexlab{a}}.

\bibitem[Du et~al.(2023{\natexlab{b}})Du, Yang, Dai, Dai, Nachum, Tenenbaum, Schuurmans, and Abbeel]{du2023learning}
Yilun Du, Mengjiao Yang, Bo~Dai, Hanjun Dai, Ofir Nachum, Joshua~B Tenenbaum, Dale Schuurmans, and Pieter Abbeel.
\newblock Learning universal policies via text-guided video generation.
\newblock \emph{arXiv e-prints}, pp.\  arXiv--2302, 2023{\natexlab{b}}.

\bibitem[Ebert et~al.(2021)Ebert, Yang, Schmeckpeper, Bucher, Georgakis, Daniilidis, Finn, and Levine]{ebert2021bridge}
Frederik Ebert, Yanlai Yang, Karl Schmeckpeper, Bernadette Bucher, Georgios Georgakis, Kostas Daniilidis, Chelsea Finn, and Sergey Levine.
\newblock Bridge data: Boosting generalization of robotic skills with cross-domain datasets.
\newblock \emph{arXiv preprint arXiv:2109.13396}, 2021.

\bibitem[Finn et~al.(2016)Finn, Goodfellow, and Levine]{finn2016unsupervised}
Chelsea Finn, Ian Goodfellow, and Sergey Levine.
\newblock Unsupervised learning for physical interaction through video prediction.
\newblock \emph{Advances in neural information processing systems}, 29, 2016.

\bibitem[Gkanatsios et~al.(2023)Gkanatsios, Jain, Xian, Zhang, Atkeson, and Fragkiadaki]{gkanatsios2023energy}
Nikolaos Gkanatsios, Ayush Jain, Zhou Xian, Yunchu Zhang, Christopher Atkeson, and Katerina Fragkiadaki.
\newblock Energy-based models as zero-shot planners for compositional scene rearrangement.
\newblock \emph{arXiv preprint arXiv:2304.14391}, 2023.

\bibitem[Grauman et~al.(2022)Grauman, Westbury, Byrne, Chavis, Furnari, Girdhar, Hamburger, Jiang, Liu, Liu, et~al.]{grauman2022ego4d}
Kristen Grauman, Andrew Westbury, Eugene Byrne, Zachary Chavis, Antonino Furnari, Rohit Girdhar, Jackson Hamburger, Hao Jiang, Miao Liu, Xingyu Liu, et~al.
\newblock Ego4d: Around the world in 3,000 hours of egocentric video.
\newblock In \emph{Proceedings of the IEEE/CVF Conference on Computer Vision and Pattern Recognition}, pp.\  18995--19012, 2022.

\bibitem[Ho et~al.(2022)Ho, Chan, Saharia, Whang, Gao, Gritsenko, Kingma, Poole, Norouzi, Fleet, et~al.]{ho2022imagen}
Jonathan Ho, William Chan, Chitwan Saharia, Jay Whang, Ruiqi Gao, Alexey Gritsenko, Diederik~P Kingma, Ben Poole, Mohammad Norouzi, David~J Fleet, et~al.
\newblock Imagen video: High definition video generation with diffusion models.
\newblock \emph{arXiv preprint arXiv:2210.02303}, 2022.

\bibitem[Huang et~al.(2022{\natexlab{a}})Huang, Abbeel, Pathak, and Mordatch]{huang2022language}
Wenlong Huang, Pieter Abbeel, Deepak Pathak, and Igor Mordatch.
\newblock Language models as zero-shot planners: Extracting actionable knowledge for embodied agents.
\newblock In \emph{International Conference on Machine Learning}, pp.\  9118--9147. PMLR, 2022{\natexlab{a}}.

\bibitem[Huang et~al.(2022{\natexlab{b}})Huang, Xia, Xiao, Chan, Liang, Florence, Zeng, Tompson, Mordatch, Chebotar, et~al.]{huang2022inner}
Wenlong Huang, Fei Xia, Ted Xiao, Harris Chan, Jacky Liang, Pete Florence, Andy Zeng, Jonathan Tompson, Igor Mordatch, Yevgen Chebotar, et~al.
\newblock Inner monologue: Embodied reasoning through planning with language models.
\newblock \emph{arXiv preprint arXiv:2207.05608}, 2022{\natexlab{b}}.

\bibitem[Huang et~al.(2023)Huang, Xia, Shah, Driess, Zeng, Lu, Florence, Mordatch, Levine, Hausman, et~al.]{huang2023grounded}
Wenlong Huang, Fei Xia, Dhruv Shah, Danny Driess, Andy Zeng, Yao Lu, Pete Florence, Igor Mordatch, Sergey Levine, Karol Hausman, et~al.
\newblock Grounded decoding: Guiding text generation with grounded models for robot control.
\newblock \emph{arXiv preprint arXiv:2303.00855}, 2023.

\bibitem[Kaelbling \& Lozano-P{\'e}rez(2011)Kaelbling and Lozano-P{\'e}rez]{kaelbling2011hierarchical}
Leslie~Pack Kaelbling and Tom{\'a}s Lozano-P{\'e}rez.
\newblock Hierarchical task and motion planning in the now.
\newblock In \emph{2011 IEEE International Conference on Robotics and Automation}, pp.\  1470--1477. IEEE, 2011.

\bibitem[Kwon \& Han(2005)Kwon and Han]{kwon2005receding}
Wook~Hyun Kwon and Soo~Hee Han.
\newblock \emph{Receding horizon control: model predictive control for state models}.
\newblock Springer Science \& Business Media, 2005.

\bibitem[Li et~al.(2022)Li, Puig, Paxton, Du, Wang, Fan, Chen, Huang, Aky{\"u}rek, Anandkumar, et~al.]{li2022pre}
Shuang Li, Xavier Puig, Chris Paxton, Yilun Du, Clinton Wang, Linxi Fan, Tao Chen, De-An Huang, Ekin Aky{\"u}rek, Anima Anandkumar, et~al.
\newblock Pre-trained language models for interactive decision-making.
\newblock \emph{Advances in Neural Information Processing Systems}, 35:\penalty0 31199--31212, 2022.

\bibitem[Liang et~al.(2023)Liang, Huang, Xia, Xu, Hausman, Ichter, Florence, and Zeng]{liang2023code}
Jacky Liang, Wenlong Huang, Fei Xia, Peng Xu, Karol Hausman, Brian Ichter, Pete Florence, and Andy Zeng.
\newblock Code as policies: Language model programs for embodied control.
\newblock In \emph{2023 IEEE International Conference on Robotics and Automation (ICRA)}, pp.\  9493--9500. IEEE, 2023.

\bibitem[Liu et~al.(2021)Liu, Li, Du, Tenenbaum, and Torralba]{liu2021learning}
Nan Liu, Shuang Li, Yilun Du, Josh Tenenbaum, and Antonio Torralba.
\newblock Learning to compose visual relations.
\newblock \emph{Advances in Neural Information Processing Systems}, 34:\penalty0 23166--23178, 2021.

\bibitem[Liu et~al.(2022)Liu, Li, Du, Torralba, and Tenenbaum]{liu2022compositional}
Nan Liu, Shuang Li, Yilun Du, Antonio Torralba, and Joshua~B Tenenbaum.
\newblock Compositional visual generation with composable diffusion models.
\newblock \emph{arXiv preprint arXiv:2206.01714}, 2022.

\bibitem[Lynch et~al.(2023)Lynch, Wahid, Tompson, Ding, Betker, Baruch, Armstrong, and Florence]{lynch2023interactive}
Corey Lynch, Ayzaan Wahid, Jonathan Tompson, Tianli Ding, James Betker, Robert Baruch, Travis Armstrong, and Pete Florence.
\newblock Interactive language: Talking to robots in real time.
\newblock \emph{IEEE Robotics and Automation Letters}, 2023.

\bibitem[Nie et~al.(2021)Nie, Vahdat, and Anandkumar]{lace}
Weili Nie, Arash Vahdat, and Anima Anandkumar.
\newblock Controllable and compositional generation with latent-space energy-based models.
\newblock \emph{Advances in Neural Information Processing Systems}, 34, 2021.

\bibitem[Oh et~al.(2015)Oh, Guo, Lee, Lewis, and Singh]{oh2015action}
Junhyuk Oh, Xiaoxiao Guo, Honglak Lee, Richard~L Lewis, and Satinder Singh.
\newblock Action-conditional video prediction using deep networks in atari games.
\newblock \emph{Advances in neural information processing systems}, 28, 2015.

\bibitem[Oprea et~al.(2020)Oprea, Martinez-Gonzalez, Garcia-Garcia, Castro-Vargas, Orts-Escolano, Garcia-Rodriguez, and Argyros]{oprea2020review}
Sergiu Oprea, Pablo Martinez-Gonzalez, Alberto Garcia-Garcia, John~Alejandro Castro-Vargas, Sergio Orts-Escolano, Jose Garcia-Rodriguez, and Antonis Argyros.
\newblock A review on deep learning techniques for video prediction.
\newblock \emph{IEEE Transactions on Pattern Analysis and Machine Intelligence}, 44\penalty0 (6):\penalty0 2806--2826, 2020.

\bibitem[Po \& Wetzstein(2023)Po and Wetzstein]{po2023compositional}
Ryan Po and Gordon Wetzstein.
\newblock Compositional 3d scene generation using locally conditioned diffusion.
\newblock \emph{arXiv preprint arXiv:2303.12218}, 2023.

\bibitem[Saxena et~al.(2021)Saxena, Ba, and Hafner]{saxena2021clockwork}
Vaibhav Saxena, Jimmy Ba, and Danijar Hafner.
\newblock Clockwork variational autoencoders.
\newblock \emph{Advances in Neural Information Processing Systems}, 34:\penalty0 29246--29257, 2021.

\bibitem[Schuhmann et~al.(2022)Schuhmann, Beaumont, Vencu, Gordon, Wightman, Cherti, Coombes, Katta, Mullis, Wortsman, et~al.]{schuhmann2022laion}
Christoph Schuhmann, Romain Beaumont, Richard Vencu, Cade Gordon, Ross Wightman, Mehdi Cherti, Theo Coombes, Aarush Katta, Clayton Mullis, Mitchell Wortsman, et~al.
\newblock Laion-5b: An open large-scale dataset for training next generation image-text models.
\newblock \emph{Advances in Neural Information Processing Systems}, 35:\penalty0 25278--25294, 2022.

\bibitem[Selman \& Gomes(2006)Selman and Gomes]{selman2006hill}
Bart Selman and Carla~P Gomes.
\newblock Hill-climbing search.
\newblock \emph{Encyclopedia of cognitive science}, 81:\penalty0 82, 2006.

\bibitem[Tellex et~al.(2020)Tellex, Gopalan, Kress-Gazit, and Matuszek]{tellex2020robots}
Stefanie Tellex, Nakul Gopalan, Hadas Kress-Gazit, and Cynthia Matuszek.
\newblock Robots that use language.
\newblock \emph{Annual Review of Control, Robotics, and Autonomous Systems}, 3:\penalty0 25--55, 2020.

\bibitem[Urain et~al.(2021)Urain, Li, Liu, D'Eramo, and Peters]{urain2021composable}
Julen Urain, Anqi Li, Puze Liu, Carlo D'Eramo, and Jan Peters.
\newblock Composable energy policies for reactive motion generation and reinforcement learning.
\newblock \emph{arXiv preprint arXiv:2105.04962}, 2021.

\bibitem[Villegas et~al.(2022)Villegas, Babaeizadeh, Kindermans, Moraldo, Zhang, Saffar, Castro, Kunze, and Erhan]{villegas2022phenaki}
Ruben Villegas, Mohammad Babaeizadeh, Pieter-Jan Kindermans, Hernan Moraldo, Han Zhang, Mohammad~Taghi Saffar, Santiago Castro, Julius Kunze, and Dumitru Erhan.
\newblock Phenaki: Variable length video generation from open domain textual description.
\newblock \emph{arXiv preprint arXiv:2210.02399}, 2022.

\bibitem[Wang et~al.(2023{\natexlab{a}})Wang, Cai, Liu, Ma, and Liang]{wang2023describe}
Zihao Wang, Shaofei Cai, Anji Liu, Xiaojian Ma, and Yitao Liang.
\newblock Describe, explain, plan and select: Interactive planning with large language models enables open-world multi-task agents.
\newblock \emph{arXiv preprint arXiv:2302.01560}, 2023{\natexlab{a}}.

\bibitem[Wang et~al.(2023{\natexlab{b}})Wang, Gui, Negrea, and Veitch]{wang2023concept}
Zihao Wang, Lin Gui, Jeffrey Negrea, and Victor Veitch.
\newblock Concept algebra for text-controlled vision models.
\newblock \emph{arXiv preprint arXiv:2302.03693}, 2023{\natexlab{b}}.

\bibitem[Wolfe et~al.(2010)Wolfe, Marthi, and Russell]{wolfe2010combined}
Jason Wolfe, Bhaskara Marthi, and Stuart Russell.
\newblock Combined task and motion planning for mobile manipulation.
\newblock In \emph{Proceedings of the International Conference on Automated Planning and Scheduling}, volume~20, pp.\  254--257, 2010.

\bibitem[Wu et~al.(2022)Wu, Tjandrasuwita, Wu, Yang, Liu, Sosic, and Leskovec]{wu2022zeroc}
Tailin Wu, Megan Tjandrasuwita, Zhengxuan Wu, Xuelin Yang, Kevin Liu, Rok Sosic, and Jure Leskovec.
\newblock Zeroc: A neuro-symbolic model for zero-shot concept recognition and acquisition at inference time.
\newblock \emph{Advances in Neural Information Processing Systems}, 35:\penalty0 9828--9840, 2022.

\bibitem[Xue et~al.(2016)Xue, Wu, Bouman, and Freeman]{xue2016visual}
Tianfan Xue, Jiajun Wu, Katherine Bouman, and Bill Freeman.
\newblock Visual dynamics: Probabilistic future frame synthesis via cross convolutional networks.
\newblock \emph{Advances in neural information processing systems}, 29, 2016.

\bibitem[Yang et~al.(2023{\natexlab{a}})Yang, Du, Dai, Schuurmans, Tenenbaum, and Abbeel]{yang2023probabilistic}
Mengjiao Yang, Yilun Du, Bo~Dai, Dale Schuurmans, Joshua~B Tenenbaum, and Pieter Abbeel.
\newblock Probabilistic adaptation of text-to-video models.
\newblock \emph{arXiv preprint arXiv:2306.01872}, 2023{\natexlab{a}}.

\bibitem[Yang et~al.(2023{\natexlab{b}})Yang, Nachum, Du, Wei, Abbeel, and Schuurmans]{yang2023foundation}
Sherry Yang, Ofir Nachum, Yilun Du, Jason Wei, Pieter Abbeel, and Dale Schuurmans.
\newblock Foundation models for decision making: Problems, methods, and opportunities.
\newblock \emph{arXiv preprint arXiv:2303.04129}, 2023{\natexlab{b}}.

\bibitem[Yang et~al.(2023{\natexlab{c}})Yang, Mao, Du, Wu, Tenenbaum, Lozano-P{\'e}rez, and Kaelbling]{yang2023compositional}
Zhutian Yang, Jiayuan Mao, Yilun Du, Jiajun Wu, Joshua~B Tenenbaum, Tom{\'a}s Lozano-P{\'e}rez, and Leslie~Pack Kaelbling.
\newblock Compositional diffusion-based continuous constraint solvers.
\newblock \emph{arXiv preprint arXiv:2309.00966}, 2023{\natexlab{c}}.

\bibitem[Zeng et~al.(2022)Zeng, Wong, Welker, Choromanski, Tombari, Purohit, Ryoo, Sindhwani, Lee, Vanhoucke, et~al.]{zeng2022socratic}
Andy Zeng, Adrian Wong, Stefan Welker, Krzysztof Choromanski, Federico Tombari, Aveek Purohit, Michael Ryoo, Vikas Sindhwani, Johnny Lee, Vincent Vanhoucke, et~al.
\newblock Socratic models: Composing zero-shot multimodal reasoning with language.
\newblock \emph{arXiv preprint arXiv:2204.00598}, 2022.
\newblock URL \url{https://arxiv.org/abs/2204.00598}.

\bibitem[Zhao et~al.(2023)Zhao, Kumar, Levine, and Finn]{zhao2023learning}
Tony~Z Zhao, Vikash Kumar, Sergey Levine, and Chelsea Finn.
\newblock Learning fine-grained bimanual manipulation with low-cost hardware.
\newblock \emph{arXiv preprint arXiv:2304.13705}, 2023.

\end{thebibliography}
\bibliographystyle{iclr2024_conference}

\newpage
\appendix

\renewcommand{\thefigure}{\Roman{figure}}
\renewcommand{\thetable}{\Roman{table}}

\section{Appendix}
 In the Appendix, we provide additional results in \sect{sect:additional_results}. We provide evaluation details in \sect{sect:evaluation}, dataset details in \sect{sect:appendix_dataset}, training details in \sect{sect:training}, and planning details in \sect{sect:appendix_planning}.

\subsection{Additional Results}
\label{sect:additional_results}

Below, we provide additional experimental results. 

\begin{figure}[t]
    \centering
    \includegraphics[width=\linewidth]{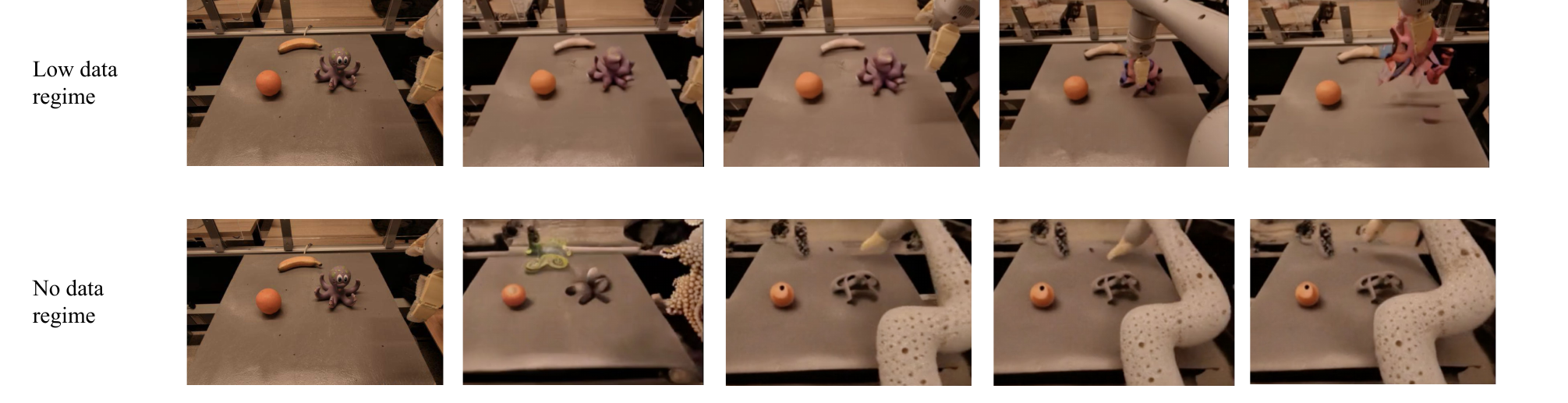}
    \vspace{-15pt}
\caption{\textbf{Failure in Transferring Web Knowledge.}  \model is instructed to ``pick  up octopus toy". In the low-data regime, the model has only seen the octopus toy in training data a few times, and the generated frames roughly reconstructed the shape of the toy, but fails at accurately recover the dynamics. The the no-data regmie, the model has never seen in training data, and without any training data \model transfers the wrong web knowledge and makes the gripper octopus arms. }
\label{fig:appendix_knowledge_failure}
\end{figure}

\begin{figure}[t]
    \centering
    \includegraphics[width=\linewidth]{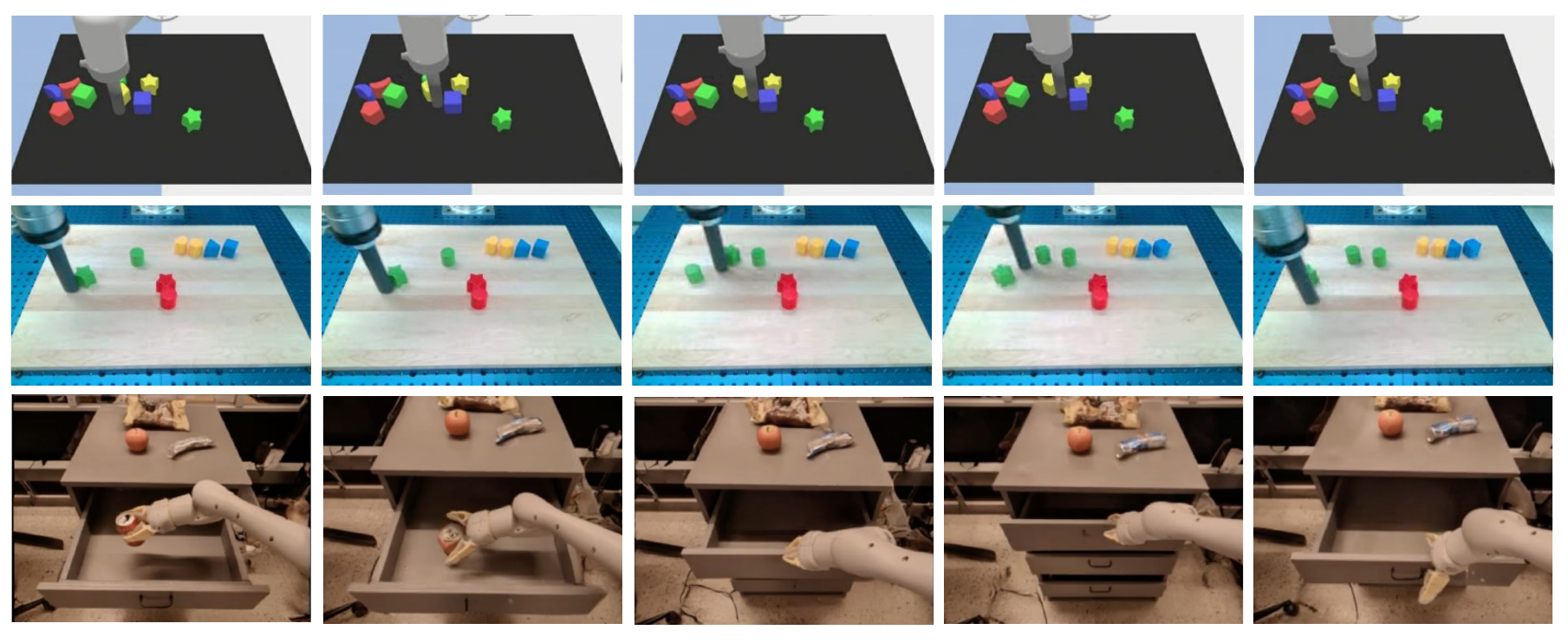}
    \vspace{-15pt}
\caption{\textbf{Failure in Physics.}  When synthesizing videos, \model will sometimes make objects dissappear or reappear. Creating long videos that respect object permanence remains an important future direction for our work.}
\label{fig:failure_physics}
\end{figure}

\myparagraph{Failure Cases.} We observed several failures cases when applying our approach across tasks. First, when using world knowledge from other text-to-video datasets to generalize and execute new tasks, we found that sometimes the video model would incorrectly interpret tasks as illustrated in \fig{fig:appendix_knowledge_failure}, where it interprets the manipulator as an octopus. In addition, we found that our approach sometimes synthesized videos with inconsistent physics, where objects would infrequently appear or disappear as illustrated in \fig{fig:failure_physics}.

\begin{figure}[t]
    \centering
    \includegraphics[width=\linewidth]{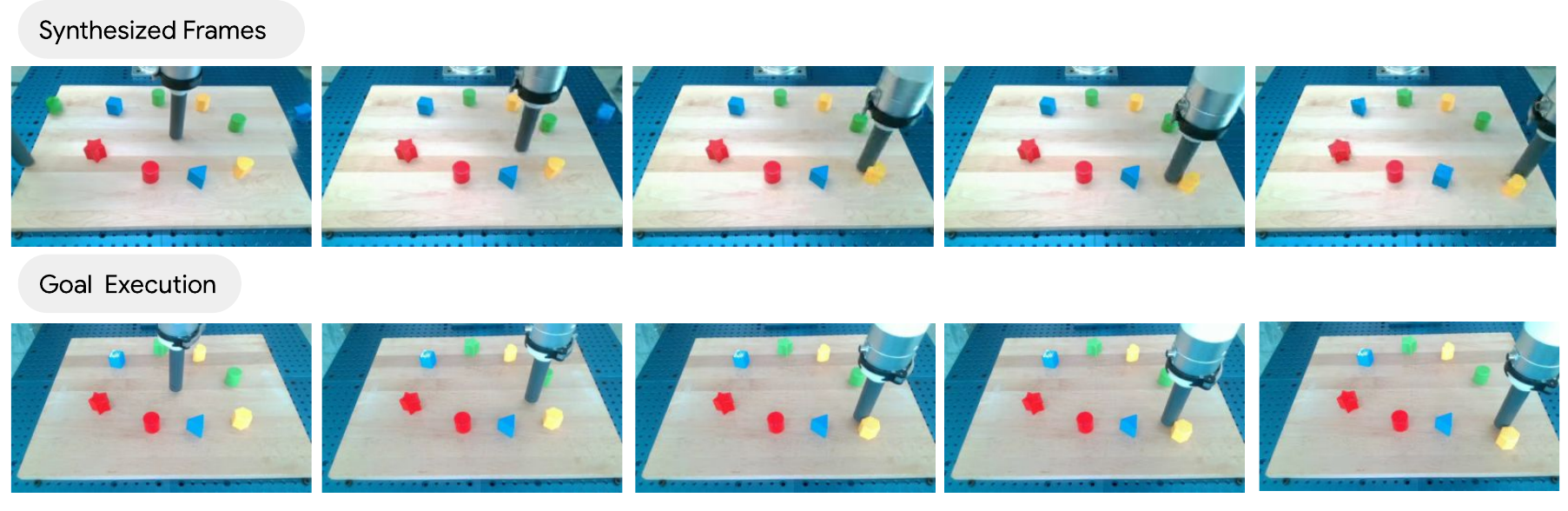}
    \vspace{-15pt}
\caption{\textbf{Goal Policy Robustness to Synthesized Goals}  The goal-conditioned policy is robust to noise in synthesized goals. In the above example, given synthesized goals in the top row, the policy can directly execute a control actions in the real environment in the bottom row. }
\label{fig:appendix_goal_robustness}
\end{figure}

\myparagraph{Robustness of Goal-Conditioned Policy} We found that our decomposition of execution in \model to visual goal generation and subsequent execution  allowed for strong generalization in the goal-conditioned policy. This was because the goal-conditioned policy can discard most information in image goals and focus only on the portion required for immediate execution of the policy. We illustrate this robustness in Fig.~\ref{fig:appendix_goal_robustness}. Even when the generated goals are misaligned with the observed board, with many artifacts in the generated boundaries, our goal-conditioned policy is still able to execute the visual task.

\begin{figure}[t]
    \centering
    \includegraphics[width=\linewidth]{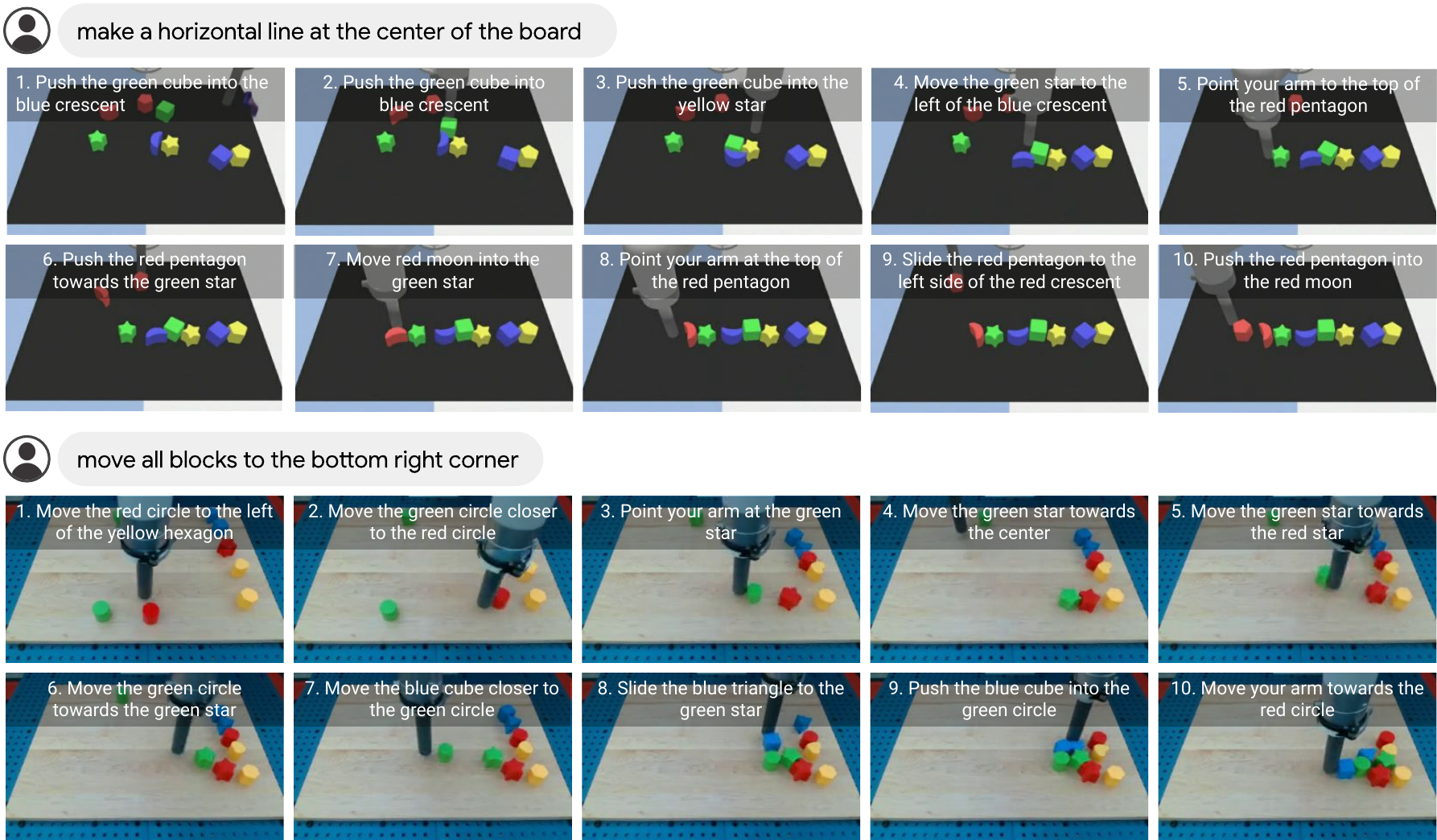}
    \vspace{-15pt}
\caption{\textbf{Additional Long Horizon Video Plans.} Additional long horizon video plans generated by \model on both simulated and real images. \model is {\it only given} the {\it initial image} and  {\it language goal}. Language subplans and other image frames are {\it directly synthesized}.}
\label{fig:appendix_plan_fig}
\end{figure}

\myparagraph{Additional Video Plans.} We provide additional video plans synthesized by our approach on block arrangement in \fig{fig:appendix_plan_fig}. Our approach is able to robustly synthesize both language and video plans.

\subsection{Evaluation Details}
\label{sect:evaluation}

\myparagraph{Video Evaluation.} To evaluate whether a generated video satisfied a long horizon task, we manually assessed whether at any point in the generated video, the image state in the video satisfied the specified long-horizon goal. Due to the computational cost of generating long-horizon video (taking approximately 30 minutes per video), we generated a total of 50 videos for each goal across each task and method.

\myparagraph{Execution Evaluation.} We used the ground truth simulation state of each block in the Language Table environment to compute rewards and task completion thresholds for each task in the paper. To compute the reward function for ``group blocks by color'', we assessed the percentage of blocks of the same color that were within 0.1 units of each other. To compute the reward for ``move blocks to an area'', we assessed the number of blocks within 0.2 x units and 0.27 y units to a specified corner.
To compute the reward for ``move blocks into a line in the center of a board'', we computed the number of blocks with 0.05 units of the center of the board. We gave all baselines and \model a total of 1500 timesteps of execution in each environment, with performance measured at the end of 1500 timesteps. If at an intermediate timestep the task was complete, we terminated execution early. 

We evaluated each method on a total of 50 environments on each task due to slowness in execution of both our method and baselines (\model took approximately 1 hour per environment, while RT2 baselines took approximate 0.5 hours per environment). For \model, we generated a plan of horizon length 2 with beam width 2 and branching factor 16, where each video in a planning horizon corresponded to 16 frames. We called the goal-conditioned policy a total of 4 times for each of the first 16 frames of the synthesized video plan in simulated environments. On real world evaluation tasks, we called the goal-conditioned policy on the first 10 frames of the synthesized video plan to more finely account for differences in physics of the real world execution compared to those used to gather the data.

\subsection{Dataset Details}
\label{sect:appendix_dataset}

\myparagraph{Language Table.} We trained \model on approximately 10000 long horizon trajectories in both simulation and real across a set of several hundred different long horizon goals. We chose a total of 3 goals to run experiments on as they allowed easy automated evaluation. These combined roughly 20000 trajectories and had approximately 400000 short-horizon text labels.

\myparagraph{7DoF Mobile Manipulator.} For our 7DoF mobile manipulator planning and execution experiments, we trained \model on the dataset from RT-1~\citep{brohan2022rt}.  For our generalization experiments of \model, we train a large text-to-video diffusion model using a mix of data from our particular 7DoF Mobile Manipulator, Bridge~\citep{ebert2021bridge}, RT-2~\citep{brohan2023rt}, Ego4D~\citep{grauman2022ego4d}, EPIC-KITCHEN~\citep{damen2018scaling}, and LAION-400M~\citep{schuhmann2022laion}.

\myparagraph{14DoF Bi-Manual Manipulation.} For our 14DoF Bi-Manual Manipulation, we train \model on approximately 1200 teleoped demonstrations on a kitchen stacking task, where operators were first asked to stack bowls on top of each other, then cups on top of bowls, and then utensils on top of cups. Each demonstration was annotated with roughly 20 language instructions, leading to roughly 25k short-horizon text labels.

\subsection{Training Details}
\label{sect:training}

\myparagraph{Video Models.} To train video diffusion models, we follow the model architecture and training approach from~\citep{du2023learning}. We train a base text-conditioned video generation model at $24\times40$ resolution and upsample it to $48\times80$ resolution and then $192\times320$ resolution, where videos at each resolution are set to generate a temporal video of length 16 frames. We use a base channel width of 256 across models and train a base text-conditioned video model using 64 TPUv3 pods for 3 days and higher resolution superresolution models for 1 day. We train separate text-to-video models per domain.

\myparagraph{VLM Models.} To train VLMs, we follow the architecture and codebase of PaLM-E~\citep{driess2023palm}. We fine-tune a single 12B PaLM-E~\citep{driess2023palm} jointly to both predict heuristics and policies. We finetune the VLM model using 64 TPUv3 pods for 1 day on data in each domain and use separate models per domain.

\myparagraph{Goal-Conditioned Policy.} To train our goal-conditioned policy, we use the LAVA model~\citep{lynch2023interactive} architecture, where the text-encoder from CLIP is replaced with a ResNet encoder for the goal image. We train our goal-conditioned policy in each domain using 16 TPUv3 pods for 1 day. 

\myparagraph{Baselines.} We compare our approach with a set of multiple baselines. For the LAVA baseline, we follow the exact settings from the LAVA paper ~\citep{lynch2023interactive}, where we train the model conditioned on long-horizon goal text and trained the model for 1 day use 16 TPUv3 pods. For our UniPi~\citep{du2023learning} baseline, we followed the same training architecture and approach as the text-conditioned video model in \model, where the model is trained on long-horizon video elements. For the RT-2 baseline, we follow the architecture and codebase of the original paper~\citep{brohan2023rt}, where we use the 12B RT2-PaLM-E model. We trained RT-2 using 64 TPUv3 pods for 3 days. 

\subsection{Planning Details}
\label{sect:appendix_planning}

\myparagraph{Language Table.} To generate video plans, we planned with a horizon of 16, a beam width of 2, a language branching factor of 4, and a video branching factor of 4. To enable fast video generation, we used the DDIM sampler, with a total of 64 timesteps of sampling at the base resolution and 4 timesteps of sampling at the higher resolution samples, with a classifier-free guidance scale of 5 for the base model. We generated 16 videos in parallel at once use a 4 TPU inference pod.

We queried the VLM policy to generate different text actions given an image with a temperature of 0.3. Our VLM heuristic function decoded the number of steps left until task-completion with a temperature of 0.0. We set our heuristic function clipping threshold during planning to be 50 and removed videos if the improvement was larger than 50 after one video rollout. This number was heuristically determined based off generated plans at different thresholds, with the aim of finding the highest threshold such that long-horizon plans still looked physically plausible.  Our total planning procedure across a horizon of 16 took approximately 30 minutes.

\myparagraph{7DoF Mobile Manipulator.} In the 7DoF mobile manipulator experiments we made slight tweaks to the approach. For VLM Planning, we used PaLM-E~(\cite{driess2023palm}) to generate scene captions, and few-shot prompted PaLM~(\cite{chowdhery2022palm}) to generate plans following the prompts in SayCan~(\cite{ahn2022can}). The beam search has a beam width of 3. The video diffusion model is the same as above, except that it has a different resolution for the base model ($64\times80$) and super resolution model ($256\times320$), and is finetuned on RT-1~(\cite{brohan2022rt}) data only. The goal-conditioned policy is using the last frame of the generated video segment only.

\myparagraph{14DoF Bi-Manual manipulation.} We followed the same planning setup as in Language Table. We set our heuristic function clipped threshold during planning to be 15.  


\end{document}